\documentclass[10pt,twocolumn,letterpaper]{article}
\pdfoutput=1

\usepackage[pagenumbers]{cvpr} 

\usepackage{graphicx}
\usepackage{amsmath}
\usepackage{amssymb}
\usepackage{booktabs}
\usepackage{xcolor}
\usepackage{multirow}
\usepackage{pifont}
\usepackage{color, colortbl}
\usepackage[normalem]{ulem}
\usepackage{hhline}
\usepackage{subcaption, siunitx}
\usepackage{makecell}
\usepackage{placeins}
\usepackage[accsupp]{axessibility}
\usepackage{multicol}

\newcommand{\cmark}{\ding{51}}%
\newcommand{\xmark}{\ding{55}}%

\newcommand*{\MySmallestIndent}{\hspace*{-0.1cm}}
\newcommand*{\MySmallestIndentRight}{\hspace*{0.1cm}}
\definecolor{Gray}{gray}{0.9}
\newcommand{\udensdash}[1]{\dashuline{#1}}%

\usepackage[pagebackref,breaklinks,colorlinks]{hyperref}

\usepackage[capitalize]{cleveref}
\crefname{section}{Sec.}{Secs.}
\Crefname{section}{Section}{Sections}
\Crefname{table}{Table}{Tables}
\crefname{table}{Tab.}{Tabs.}

\def\cvprPaperID{9421} 
\def\confName{CVPR}
\def\confYear{2023}

\begin{document}

\title{Interactive and Explainable Region-guided Radiology Report Generation}

\author{Tim Tanida\textsuperscript{1,}\thanks{Equal contribution} \qquad
    Philip Müller\textsuperscript{1,}\footnotemark[1] \qquad
    Georgios Kaissis\textsuperscript{1,2} \qquad Daniel Rueckert\textsuperscript{1,3} \\
    \textsuperscript{1}Technical University of Munich, \textsuperscript{2}Helmholtz Zentrum Munich, \textsuperscript{3}Imperial College London \\
    {\tt\small \{tim.tanida, philip.j.mueller, g.kaissis, daniel.rueckert\}@tum.de}
}
\maketitle

\begin{abstract}
The automatic generation of radiology reports has the potential to assist radiologists in the time-consuming task of report writing. Existing methods generate the full report from image-level features, failing to explicitly focus on anatomical regions in the image. We propose a simple yet effective region-guided report generation model that detects anatomical regions and then describes individual, salient regions to form the final report. While previous methods generate reports without the possibility of human intervention and with limited explainability, our method opens up novel clinical use cases through additional interactive capabilities and introduces a high degree of transparency and explainability. Comprehensive experiments demonstrate our method's effectiveness in report generation, outperforming previous state-of-the-art models, and highlight its interactive capabilities. The code and checkpoints are available at \url{https://github.com/ttanida/rgrg} .
\end{abstract}

\section{Introduction}
\label{introduction}

Chest radiography (chest X-ray) is the most common type of medical image examination in the world and is critical for identifying common thoracic diseases such as pneumonia and lung cancer \cite{raoof2012interpretation, johnson2019mimicphysio}. Given a chest X-ray, radiologists examine each depicted anatomical region and describe findings of both normal and abnormal salient regions in a textual report\cite{goergen2013evidence}. Given the large volume of chest X-rays to be examined in daily clinical practice, this often becomes a time-consuming and difficult task, which is further exacerbated by a shortage of trained radiologists in many healthcare systems \cite{rosenkrantz2016us, rimmer2017radiologist, bastawrous2017improving}. As a result, automatic radiology report generation has emerged as an active research area with the potential to alleviate radiologists' workload.

\begin{figure}[t]
  \centering
   \includegraphics[clip, trim=0.5cm 12.8cm 0.5cm 9cm, width=1.0\linewidth]{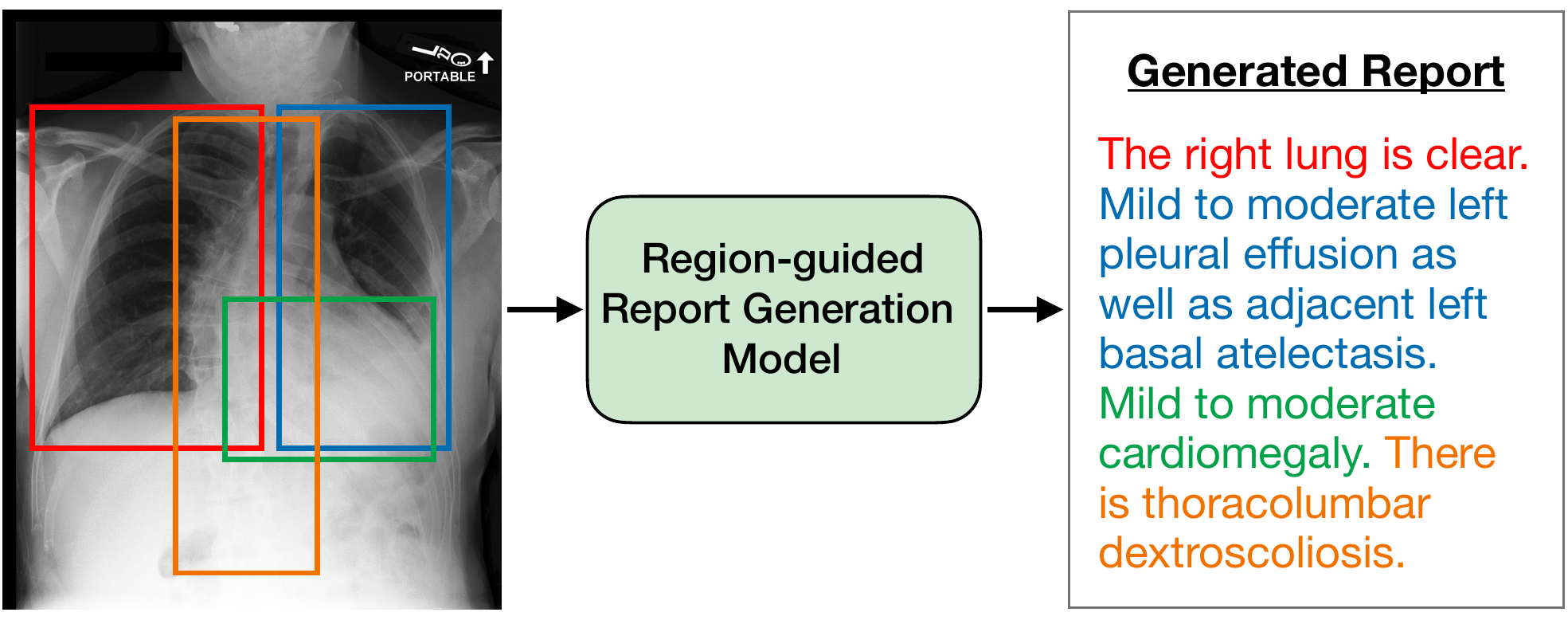}
   \caption{Our approach at a glance. Unique anatomical regions of the chest are detected, the most salient regions are selected for the report and individual sentences are generated for each region. Consequently, each sentence in the generated report is explicitly grounded on an anatomical region.}
   \label{fig:high-level-overview}
\end{figure}

Generating radiology reports is a difficult task since reports consist of multiple sentences, each describing a specific medical observation of a specific anatomical region. As such, current radiology report generation methods tend to generate reports that are factually incomplete (\ie, missing key observations in the image) and inconsistent (\ie, containing factually wrong information) \cite{miura2021improving}. This is further exacerbated by current methods utilizing image-level visual features to generate reports, failing to explicitly focus on salient anatomical regions in the image. Another issue regarding existing methods is the lack of explainability. A highly accurate yet opaque report generation system may not achieve adoption in the safety-critical medical domain if the rationale behind a generated report is not transparent and explainable \cite{guidotti2018survey, miller2019explanation, geis2019ethics}. 
Lastly, current methods lack interactivity and adaptability to radiologists' preferences. \Eg, a radiologist may want a model to focus exclusively on specific anatomical regions within an image.

Inspired by radiologists' working patterns, we propose a simple yet effective \textbf{R}egion-\textbf{G}uided Radiology \textbf{R}eport \textbf{G}eneration (RGRG) method to address the challenges highlighted before. Instead of relying on image-level visual features, our work is the first in using object detection to directly extract localized visual features of anatomical regions, which are used to generate individual, anatomy-specific sentences describing any pathologies to form the final report. Conceptually, we \textit{divide and conquer} the difficult task of generating a complete and consistent report from the whole image into a series of simple tasks of generating short, consistent sentences for a range of isolated anatomical regions, thereby achieving both completeness and consistency for the final, composed report.

While existing models \cite{jing2018automatic, liu2019clinically, chen2020generating} can produce heatmaps to illustrate which parts of an image were attended during report generation, our model visually grounds each sentence in the generated report on a predicted bounding box around an anatomical region (see \cref{fig:high-level-overview}), thereby introducing a high degree of explainability to the report generation process. The visual grounding allows radiologists to easily verify the correctness of each generated sentence, which may increase trust in the model's predictions \cite{reyes2020interpretability, miller2019explanation, geis2019ethics}. 
Moreover, our approach enables interactive use, allowing radiologists to select anatomical structures or draw bounding boxes around regions of interest for targeted description generation, enhancing flexibility in the clinical workflow.

Our contributions are as follows:
\begin{itemize}
  \item We introduce a simple yet effective \textbf{R}egion-\textbf{G}uided Radiology \textbf{R}eport \textbf{G}eneration (RGRG) method that detects anatomical regions and generates individual descriptions for each. Empirical evidence demonstrates that our model produces relevant sentences pertaining to the anatomy. To the best of our knowledge, it is the first report generation model to visually ground sentences to anatomical structures.
  \item We condition a pre-trained language model on each anatomical region independently. This enables \textbf{anatomy-based sentence generation}, where radiologists interactively select individual, detected anatomical regions for which descriptions are generated.
  \item Additionally, our approach enables radiologists to manually define regions of interest using bounding boxes, generating corresponding descriptions. We assess the impact of these manually drawn boxes, demonstrating the robustness of the \textbf{selection-based sentence generation} task.
  \item For \textbf{full radiology report generation}, a module is introduced to select salient anatomical regions, whose generated descriptions are concatenated to form factually complete and consistent reports. We empirically demonstrate this on the MIMIC-CXR dataset \cite{johnson2019mimic,johnson2019mimicphysio}, where we outperform competitive baselines in both language generation and clinically relevant metrics.
\end{itemize}

\section{Related work}
\subsection{Radiology report generation}

\noindent\textbf{Transformer-based radiology report generation.} Automatic radiology report generation has garnered significant research interest in recent years. While early works \cite{wang2018tienet, jing2018automatic, li2018hybrid, jing2019show, zhang2020radiology} adopted CNN-RNN architectures, more recent studies have shifted towards utilizing the effectiveness of the Transformer \cite{vaswani2017attention}. The standard transformer architecture was adapted by adding relation memory units \cite{chen2020generating} or a memory matrix \cite{chen2021cross} to improve cross-model interactions between visual and textual features. 
To better attend to abnormal regions, a transformer was proposed that iteratively aligns extracted visual features and disease tags \cite{you2021aligntransformer}, a system contrasting normal and abnormal images was presented \cite{liu2021contrastive}, and an additional medical knowledge graph was utilized \cite{liu2021exploring}. In contrast, we focus on abnormal regions by directly extracting the corresponding region visual features via object detection, additionally encoding strong abnormal information in the features with an abnormality classification module. 
Recently, warm-starting components from pre-trained models has shown promising results for report generation \cite{nicolson2022improving}. 
Inspired by \cite{alfarghaly2021automated}, our work uses a pre-trained language model as the decoder.

\noindent\textbf{Consecutive radiology report generation.} Similar to our approach, a variety of previous works have attempted to divide the difficult task of generating a long, coherent radiology report into several steps. Hierarchical approaches were proposed \cite{liu2019clinically, nooralahzadeh2021progressive}, in which high-level concepts are first extracted from the image and subsequently decoded into individual sentences. Our work is most related to \cite{wang2022inclusive}. They introduce a multi-head transformer applied to patch features from a CNN backbone, with each head assigned to a specific anatomical region, generating sentences exclusively for that region.
In contrast, our method extracts region-level features through object detection (as opposed to image-level features), and generates region-level sentences with a shared transformer decoder based on these features (instead of selectively using designated heads).
While our method requires more supervision, it is overall less technically complex, whilst offering additional interactivity and a higher degree of transparency and explainability.

\subsection{Image captioning}

\noindent\textbf{Image captioning.} Most radiology report generation methods \cite{alfarghaly2021automated, chen2020generating, liu2021exploring, you2021aligntransformer, liu2021contrastive, chen2021cross, jing2018automatic, wang2018tienet} are inspired by influential works \cite{xu2015show, vinyals2015show, you2016image, cornia2020meshed} from the image captioning domain in computer vision. While overarching concepts can be translated from general image captioning to radiology report generation, there are notable differences: 1) Radiology reports are much longer and more diverse than typical captions, describing multiple anatomical regions in an image. 2) Generating descriptions of specific but crucial abnormalities is complicated because of a heavy data bias towards normal images and normal reports. 

\noindent\textbf{Dense image captioning.} Our work is related to image captioning methods that utilize object detection in the encoding process, in particular works \cite{johnson2016densecap, yin2019context, li2019learning, shao2022region} from the dense image captioning domain. Instead of generating a single caption for the whole image, dense captioning methods aim to both localize and describe individual salient regions in images, usually by conditioning a language model on the specific region features. Intuitively, dense image captioning aligns more closely with radiologists' working practice, as anatomical regions in an image are usually localized and described one by one when crafting a report.

\noindent\textbf{Controllable image captioning.} In our anatomy-based sentence generation mode (see \cref{introduction}), radiologists can manually select anatomical regions automatically outlined by bounding boxes, which are then individually described by generated sentences. This is related to the domain of controllable image captioning \cite{cornia2019show, zheng2019intention, chen2020say}, in which generated captions are influenced by an external control signal, \eg by specifying the image regions to be described \cite{cornia2019show, zheng2019intention}, or by specifying an abstract scene graph \cite{chen2020say}. However, while controllable image captioning methods typically generate a single image-level caption, our model in this mode generates descriptions independently for each selected region.

\begin{figure*}[ht]
  \centering
   \includegraphics[clip, trim=0cm 2.8cm 0.25cm 2.6cm, width=1.0\textwidth]{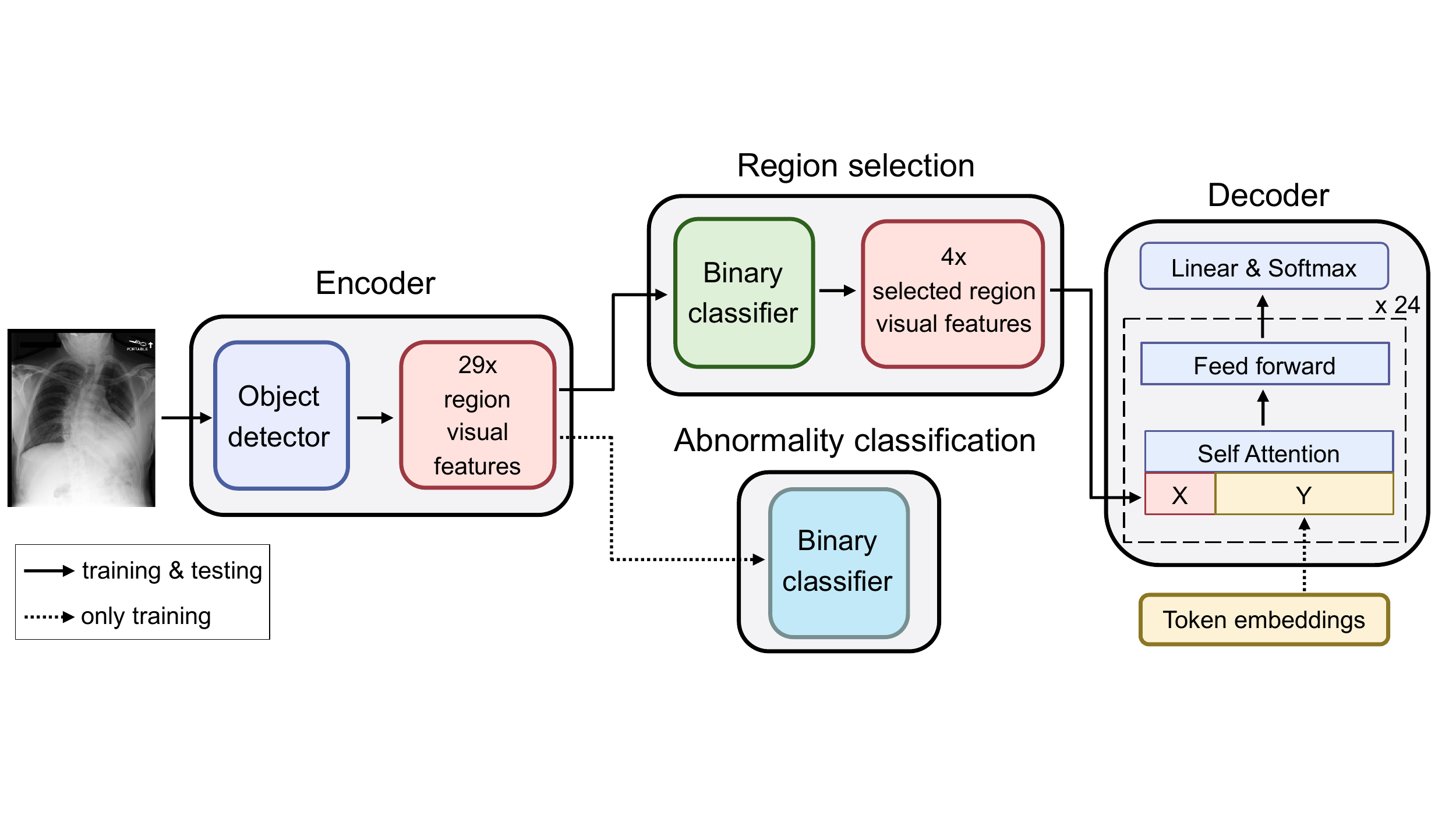}
   \caption{\textbf{R}egion-\textbf{G}uided Radiology \textbf{R}eport \textbf{G}eneration (RGRG): the object detector extracts visual features for 29 unique anatomical regions of the chest. Two subsequent binary classifiers select salient region features for the final report and encode strong abnormal information in the features, respectively. The language model generates sentences for each of the selected regions (in this example 4), forming the final report. For conciseness, residual connections and layer normalizations in the language model are not depicted.}
   \label{fig:model-architecture}
\end{figure*}

\section{Method}
\label{sec:method}

Our proposed \textbf{R}egion-\textbf{G}uided Radiology \textbf{R}eport \textbf{G}eneration (RGRG) method has three distinct use cases:

\begin{enumerate}
  \item \textbf{Radiology report generation}: The proposed model generates a full radiology report for a chest X-ray image and outlines the described anatomical regions by bounding boxes in the image, as depicted in \cref{fig:high-level-overview}.
  \item \textbf{Anatomy-based sentence generation}: The proposed model outlines anatomical regions by bounding boxes. The radiologist then selects individual anatomies of interest, for which the model generates sentences.
  \item \textbf{Selection-based sentence generation}: The radiologist manually draws a bounding box around a region of interest, for which the model then generates sentences. 
\end{enumerate}

We first outline our model in the context of radiology report generation and elaborate on anatomy- and selection-based sentence generation in \cref{sec:inference}.

\subsection{Overview}

For full report generation, our model closely follows the typical workflow of a radiologist. First, a radiologist identifies distinct anatomical regions in an X-ray image. For each region, a decision is made if the region needs to be described in the report and if it is abnormal, with the latter assessment influencing the former. Finally, each selected region is described in 1-2 short sentences in the final report.

\cref{fig:model-architecture} illustrates the proposed model architecture consisting of four major modules. First, an object detector identifies and extracts visual features of 29 distinct anatomical regions of the chest. Radiologists usually only describe a handful of salient regions in a report (see the exemplary report of \cref{fig:high-level-overview}). The region selection module simulates this behavior by predicting via a binary classifier whether sentences should be generated for each region, \eg, the image of \cref{fig:model-architecture} has four regions selected for sentence generation.

During training, the region visual features are also fed into the abnormality classification module consisting of an additional binary classifier that predicts whether a region is normal or abnormal (\ie, contains a pathology). This encodes stronger abnormality information in the region visual features from the object detector, helping both the region selection module in selecting abnormal regions for sentence generation, as well as the decoder in generating sentences that capture potential pathologies of the regions.

The decoder is a transformer-based language model pre-trained on medical abstracts. It generates sentences for each selected region (treated as independent samples) by conditioning on the associated region visual features. The final report is obtained via a post-processing step which removes generated sentences that are too similar to each other and concatenates the remaining ones.

\subsection{Modules}
\label{sec:modules}

\noindent\textbf{Object detector.} For the object detector, we use Faster R-CNN \cite{ren2015faster} with a ResNet-50 \cite{he2016deep} backbone pre-trained on ImageNet \cite{deng2009imagenet}. Faster R-CNN consists of a region proposal network (RPN), which generates object proposals (\ie, bounding boxes of potential anatomical regions) based on the feature maps extracted by the backbone from the input image. A region of interest (RoI) pooling layer maps each object proposal onto the backbone feature maps, extracting small feature maps of uniform spatial extent for the proposals. These RoI feature maps are each classified into one of the 29 anatomical region classes (and the background class) following standard procedure in Faster R-CNN.

To obtain the region visual features for each of the 29 anatomical regions, we first determine the “top” object proposal 
for each region class. A “top” object proposal for a given region class is one where the class has the highest probability score (of all classes) for the given proposal, as well as the highest score for all proposals where the same region class is also the top-1 class. If a region class does not achieve the highest score for at least one proposal, it is considered an undetected class. Undetected classes are automatically deselected by the region selection module.

The region visual features $\in \mathbb{R}^{29 \times 1024}$ are then obtained by taking the 29 RoI pooling layer features maps $\in \mathbb{R}^{29 \times 2048 \times H \times W}$ corresponding to the 29 “top” object proposals, applying 2D average pooling over the spatial dimensions and reducing the dimension from 2048 to 1024 by linear transformation.

\noindent\textbf{Region selection and abnormality classification.} The binary classifiers of these modules consist of multilayer perceptrons that gradually reduce the input features to a single output logit to predict if a region is selected for sentence generation and is abnormal, respectively.

\noindent\textbf{Language model.} For the language model, we use the 355M-parameter model GPT-2 Medium \cite{radford2019language} fine-tuned on PubMed  abstracts \cite{papanikolaou2020dare}. GPT-2 is an auto-regressive neural network based on self-attention, in which tokens in a sequence are conditioned on previous tokens for text generation. This can be expressed as (ignoring scaling factors):
\begin{equation}
\text{SA}(Y) = \text{softmax}((YW_{q})(YW_{k})^{\top})(YW_{v}),
  \label{eq:self-attention}
\end{equation}
where $Y$ represents the token embeddings, and $W_{q}$, $W_{k}$, $W_{v}$ are the query, key, and value projection parameters.

To condition the language model on the region visual features, we follow \cite{alfarghaly2021automated} in using pseudo self-attention \cite{ziegler2019encoder} to directly inject the region visual features in the self-attention of the model, \ie
\begin{equation}
\hspace*{-0.2cm}
\text{PSA}(X,Y) = \text{softmax}\left((YW_{q})\begin{bmatrix}
XU_{k}\\ 
YW_{k}
\end{bmatrix}^{\top}\right)\begin{bmatrix}
XU_{v}\\ 
YW_{v}
\end{bmatrix},
  \label{eq:pseudo-self-attention}
\end{equation}
where $X$ represents the region visual features, and $U_{k}$ and $U_{v}$ are the corresponding (newly initialized) key and value projection parameters. This allows text generation conditioned on both previous tokens and region visual features.

\subsection{Training}

The model training consists of three steps. First, only the object detector is trained, then the object detector combined with the binary classifiers, and finally the full model end-to-end with all parameters left trainable.\footnote{Note that following \cite{ziegler2019encoder}, in the language model only the projection parameters tasked with injecting the region visual features are trainable.} To train the language model, we only use region visual features that have corresponding reference sentences, assuming that the region selection module correctly selects these regions (that require sentences) at test time. For regions with several sentences, the sentences are concatenated, such that the model learns to predict several sentences in such cases.

The overall training loss is defined as
\begin{equation}
\begin{aligned}
\mathcal{L} = {} & \lambda_{\text{obj}} \cdot \mathcal{L}_{\text{obj}} + \lambda_{\text{select}}\cdot\mathcal{L}_{\text{select}} \\
&+ \lambda_{\text{abnormal}}\cdot\mathcal{L}_{\text{abnormal}} + \lambda_{\text{language}}\cdot\mathcal{L}_{\text{language}} \,,
  \label{eq:loss-function}
\end{aligned}
\end{equation}
where $\mathcal{L}_{\text{obj}}$ is the Faster R-CNN object detector loss, $\mathcal{L}_{\text{select}}$ and $\mathcal{L}_{\text{abnormal}}$ are the (weighted) binary cross-entropy losses for the two binary classifiers, and $\mathcal{L}_{\text{language}}$ is the cross-entropy loss for the language model. Based on the performance on the validation set, the loss weights are set as $\lambda_{\text{obj}}$ = 1.0, $\lambda_{\text{select}}$= 5.0, $\lambda_{\text{abnormal}}$= 5.0, and $\lambda_{\text{language}}$= 2.0.

\subsection{Inference}
\label{sec:inference}

\noindent\textbf{Radiology report generation.} Reports are formed by concatenating the generated sentences of selected anatomical regions. If pathologies span several such regions or multiple anatomically similar regions (\eg, left and right lung) have no findings, then similar sentences may be generated for these regions, leading to duplicate sentences in the generated report. To remove such duplicates, we use BERTScore \cite{zhang2019bertscore} to determine the degree of similarity, always removing the shorter sentence and keeping the longer one, as longer sentences tend to contain more clinically relevant information. \Eg, the sentences \emph{"The cardiomediastinal silhouette and hilar contours are normal.”} and \emph{"The cardiomediastinal silhouette is normal.”} are quite similar, while the longer sentence contains additional, relevant information.

\noindent\textbf{Anatomy-based sentence generation.} In this mode, radiologists can manually select anatomical regions (from the 29 overall regions) for the model to examine. First, the model detects and outlines all 29 regions by bounding boxes via the object detector. Next, the radiologists select regions of interest, which are in turn (exclusively) selected by the region selection module for sentence generation.

\noindent\textbf{Selection-based sentence generation.} In this mode, radiologists can draw bounding boxes around individual anatomical regions for examination. The bounding boxes are passed through the RoI pooling layer and further transformed into region visual features (as described in \cref{sec:modules}), which are then fed into the language model for sentence generation.

\section{Experimental setup}

We evaluate our method on all three tasks: \textbf{radiology report generation}, \textbf{anatomy-based sentence generation}, and \textbf{selection-based sentence generation} (see \cref{sec:method}). We refer to \cref{appendix:exp_setup,appendix:method} for more details on the experimental setup.

\subsection{Dataset and pre-processing}

We use the Chest ImaGenome v1.0.0 \cite{wu2021chest, wu2021chestphysio, PhysioNet} dataset to train and evaluate our proposed model. It is constructed from the MIMIC-CXR \cite{johnson2019mimic, johnson2019mimicphysio} dataset, which consists of chest X-ray images with corresponding free-text radiology reports. The Chest ImaGenome dataset contains automatically constructed scene graphs for the MIMIC-CXR images. Each scene graph describes one frontal chest X-ray image and contains bounding box coordinates for 29 unique anatomical regions in the chest, as well as sentences describing each region if they exist in the corresponding radiology report. We use the official split provided by the dataset resulting in 166,512 training images, 23,952 validation images, and 47,389 test images.

All images are resized to 512x512 while preserving the original aspect ratio, padded if needed, and normalized to zero mean and unit standard deviation. Color jitter, Gaussian noise, and affine transformations are applied as image data augmentations during training. For the sentences, we remove redundant whitespaces (\ie, line breaks, etc.). For the radiology report generation task, we follow previous work \cite{liu2019clinically, boag2020baselines, miura2021improving, nicolson2022improving} in using the \emph{findings} section of the radiology reports (of the MIMIC-CXR dataset) as our reference reports. The \emph{findings} section contains the observations made by the radiologist. Following \cite{miura2021improving, nicolson2022improving}, we discard reports with empty \emph{findings} sections, resulting in 32,711 test images with corresponding reference reports. We note that no further processing is applied to the extracted reports (as opposed to \eg \cite{nicolson2022improving}).

\subsection{Evaluation metrics}
On report level, we evaluate the model on widely used natural language generation (NLG) metrics such as BLEU \cite{papineni2002bleu}, METEOR \cite{banerjee2005meteor}, ROUGE-L \cite{lin2004rouge} and CIDEr-D \cite{vedantam2015cider}, which measure the similarity between generated and reference report by counting matching n-grams (\ie, word overlap). On sentence level, we evaluate on METEOR, a metric appropriate for evaluation on both sentence- and report-level (as opposed to \eg BLEU). Since conventional NLG metrics are ill-suited to measure the clinical correctness of generated reports \cite{boag2020baselines, liu2019clinically, pino2020inspecting}, we follow \cite{liu2019clinically, chen2020generating, miura2021improving, nicolson2022improving} in additionally reporting clinical efficacy (CE) metrics. CE metrics compare generated and reference reports w.r.t. the presence status of an array of prominent clinical observations, thus capturing the diagnostic accuracy of generated reports.

\subsection{Evaluation strategy and baselines}

\begin{table*}
\centering
\resizebox{\textwidth}{!}{
\begin{tabular}{llllllllll}
\hline
Dataset                    & Method           & Year & BLEU-1 & BLEU-2 & BLEU-3 & BLEU-4 & METEOR & ROUGE-L & CIDEr\\ \hline
\multirow{12}{*}{MIMIC-CXR} & R2Gen \cite{chen2020generating}           & 2020 & 0.353  & 0.218  & 0.145  & 0.103  & 0.142  & 0.277 & 0.406\textsuperscript{\textdagger}    \\
                            & CMN \cite{chen2021cross}            & 2021 & 0.353  & 0.218  & 0.148  & 0.106  & 0.142  & 0.278 & -    \\
                           & PPKED \cite{liu2021exploring}            & 2021 & 0.360  & 0.224  & 0.149  & 0.106  & 0.149  & 0.284 & 0.237    \\
                           & $\mathcal{M}^{2}$ TR. PROGRESSIVE \cite{nooralahzadeh2021progressive}            & 2021 & \udensdash{0.378}   & 0.232  & 0.154  & 0.107  & 0.145  & 0.272 & -    \\
                           & Contrastive Attention \cite{liu2021contrastive} & 2021 & 0.350  & 0.219  & 0.152  & 0.109  & 0.151  & 0.283 & -    \\
                           & AlignTransformer \cite{you2021aligntransformer} & 2021 & \udensdash{0.378}  & \udensdash{0.235}  & \udensdash{0.156}  & 0.112  & 0.158  & 0.283 & -    \\
                           & $\mathcal{M}^{2}$ Trans w/ NLL \cite{miura2021improving} & 2021 & -  & -  & -  & 0.105  & -  & -   & 0.445    \\
                           & $\mathcal{M}^{2}$ Trans w/ NLL+BS+f\textsubscript{C\textsubscript{E}} \cite{miura2021improving} & 2021 & -  & -  & -  & 0.111  & -  & -   & 0.492    \\
                           & $\mathcal{M}^{2}$ Trans w/ NLL+BS+f\textsubscript{C\textsubscript{EN}} \cite{miura2021improving}          & 2021 & -      & -      & -      & \udensdash{0.114}  & -      & -     & \textbf{0.509}
                           \\
                           & ITA \cite{wang2022inclusive}          & 2022 & \textbf{0.395}      & \textbf{0.253}      & 0.170      & 0.121  & 0.147      & 0.284     & -
                           \\
                           & CvT-21{\large 2}DistilGPT2 \cite{nicolson2022improving}          & 2022 & 0.392      & 0.245      & 0.169      & 0.124  & 0.153      & \textbf{0.285}     & 0.361
                           \\ \cline{2-10} 
                           & RGRG       & Ours & 0.373    & 0.249    & \textbf{0.175}    & \textbf{0.126}    & \textbf{0.168}    & 0.264  & 0.495     \\ \hline
\end{tabular}}
\caption{Natural language generation (NLG) metrics for the full report generation task. Our model is competitive with or outperforms previous state-of-the-art models on a variety of metrics. Dashed lines highlight the BLEU scores of the best baselines without processed (\ie lowercased) reference reports (since lowercasing increases BLEU scores \cite{post2018call}). CIDEr score denoted by \textdagger{} cited from \cite{miura2021improving}.}
\label{tab:NLG-metrics}
\end{table*}

\begin{table*}[t!]
	\centering
	\resizebox{\textwidth}{!}{
		\begin{tabular}{lllllll||lll}
			\hline
			Dataset                    & Method                                                                                           & RL                      & Year                                    & P\textsubscript{mic-5}                              & R\textsubscript{mic-5}                              & F\textsubscript{1, mic-5}                           & P\textsubscript{ex-14}                          & R\textsubscript{ex-14}                          & F\textsubscript{1, ex-14}                       \\ \hline
			\multirow{9}{*}{MIMIC-CXR} & \cellcolor{Gray} R2Gen \cite{chen2020generating}                                                 & \MySmallestIndentRight\cellcolor{Gray}\xmark  & \MySmallestIndent \cellcolor{Gray} 2020 & \MySmallestIndent\cellcolor{Gray} 0.412             & \MySmallestIndent\cellcolor{Gray} 0.298             & \MySmallestIndent\cellcolor{Gray} 0.346             & \cellcolor{Gray}0.331          & \cellcolor{Gray}0.224          & \cellcolor{Gray}0.228          \\
			                           & \cellcolor{Gray} $\mathcal{M}^{2}$ Trans w/ NLL \cite{miura2021improving}                        & \cellcolor{Gray} \xmark & \MySmallestIndent\cellcolor{Gray} 2021  & \MySmallestIndent\cellcolor{Gray} \udensdash{0.489} & \MySmallestIndent\cellcolor{Gray} \udensdash{0.411} & \MySmallestIndent\cellcolor{Gray} \udensdash{0.447} & \cellcolor{Gray}-              & \cellcolor{Gray}-              & \cellcolor{Gray}-              \\
			                           & \MySmallestIndentRight$\mathcal{M}^{2}$ Trans w/ NLL+BS+f\textsubscript{C\textsubscript{E}} \cite{miura2021improving}  & \MySmallestIndentRight\cmark                  & 2021                                    & 0.463                                               & \textbf{0.732}                                      & \textbf{0.567}                                      & -                                               & -                                               & -                                               \\
			                           & \MySmallestIndentRight$\mathcal{M}^{2}$ Trans w/ NLL+BS+f\textsubscript{C\textsubscript{EN}} \cite{miura2021improving} & \MySmallestIndentRight\cmark                  & 2021                                    & \textbf{0.503}                                      & 0.651                                               & \textbf{0.567}                                      & -                                               & -                                               & -                                               \\
			                           & \cellcolor{Gray} CMN \cite{chen2021cross}                                                        & \cellcolor{Gray}\MySmallestIndentRight\xmark  & \MySmallestIndent \cellcolor{Gray} 2021 & \MySmallestIndent\cellcolor{Gray} -                 & \MySmallestIndent\cellcolor{Gray} -                 & \MySmallestIndent\cellcolor{Gray} -                 & \cellcolor{Gray}0.334          & \cellcolor{Gray}0.275          & \cellcolor{Gray}0.278          \\
			                           & \cellcolor{Gray} Contrastive Attention \cite{liu2021contrastive}                                 & \cellcolor{Gray}\MySmallestIndentRight\xmark  & \MySmallestIndent \cellcolor{Gray} 2021 & \MySmallestIndent\cellcolor{Gray} -                 & \MySmallestIndent\cellcolor{Gray} -                 & \MySmallestIndent\cellcolor{Gray} -                 & \cellcolor{Gray}0.352          & \cellcolor{Gray}0.298          & \cellcolor{Gray}0.303          \\
			                           & \cellcolor{Gray} $\mathcal{M}^{2}$ TR. PROGRESSIVE \cite{nooralahzadeh2021progressive}           & \cellcolor{Gray}\MySmallestIndentRight\xmark  & \MySmallestIndent \cellcolor{Gray} 2021 & \MySmallestIndent\cellcolor{Gray} -                 & \MySmallestIndent\cellcolor{Gray} -                 & \MySmallestIndent\cellcolor{Gray} -                 & \cellcolor{Gray}0.240          & \cellcolor{Gray}0.428          & \cellcolor{Gray}0.308          \\
			                           & \cellcolor{Gray} CvT-21{\large 2}DistilGPT2 \cite{nicolson2022improving}                         & \cellcolor{Gray}\MySmallestIndentRight\xmark  & \MySmallestIndent \cellcolor{Gray} 2022 & \MySmallestIndent\cellcolor{Gray} -                 & \MySmallestIndent\cellcolor{Gray} -                 & \MySmallestIndent\cellcolor{Gray} -                 & \cellcolor{Gray}0.359          & \cellcolor{Gray}0.412          & \cellcolor{Gray}0.384          \\
			\hhline{~|*9{-}|}
			                           & \cellcolor{Gray}RGRG                                                                             & \cellcolor{Gray}\MySmallestIndentRight\xmark  & \MySmallestIndent\cellcolor{Gray} Ours  & \MySmallestIndent\cellcolor{Gray} \underline{0.491} & \MySmallestIndent\cellcolor{Gray} \underline{0.617} & \MySmallestIndent\cellcolor{Gray} \underline{0.547} & \cellcolor{Gray}\textbf{0.461} & \cellcolor{Gray}\textbf{0.475} & \cellcolor{Gray}\textbf{0.447} \\ \hline
		\end{tabular}
	}
	\caption{Clinical efficacy (CE) metrics micro-averaged over 5 observations (denoted by mic-5) and example-based averaged over 14 observations (denoted by ex-14). RL represents reinforcement learning. Our model outperforms all non-RL models by large margins and is competitive with the two RL-based models directly optimized on CE metrics. Dashed lines highlight the scores of the best non-RL baseline. Micro-averaged results of R2Gen cited from \cite{miura2021improving}, all example-based results are cited from \cite{nicolson2022improving}.}
	\label{tab:CE-metrics-total}
\end{table*}

\noindent\textbf{Radiology report generation.} We compare our model with previous state-of-the-art models, specifically R2Gen \cite{chen2020generating}, CMN \cite{chen2021cross}, PPKED \cite{liu2021exploring}, $\mathcal{M}^{2}$ TR. PROGRESSIVE \cite{nooralahzadeh2021progressive}, Contrastive Attention \cite{liu2021contrastive}, AlignTransformer \cite{you2021aligntransformer}, $\mathcal{M}^{2}$ Trans \cite{cornia2020meshed} optimized on a standard language model loss and rewards towards factual completeness and consistency \cite{miura2021improving}, ITA \cite{wang2022inclusive}, CvT-21{\large 2}DistilGPT2 \cite{nicolson2022improving}. If not denoted otherwise, we cite the results from the corresponding papers.

\noindent\textbf{Anatomy-based sentence generation.} To the best of our knowledge, this is the first work in using a model capable of generating sentences for specific anatomical regions of a chest X-ray. As such, we have no direct baselines to compare against and therefore conduct a qualitative analysis. Additionally, we report the performance of the language model in this task. Specifically, we report the per-anatomy METEOR scores (\ie, computed between the generated and reference sentences of each anatomy independently) for six prominent regions as well as the micro-average over all, normal, and abnormal regions, respectively. Additionally, for further verification of the model generating anatomy-sensitive sentences, we compute a custom "Anatomy-Sensitivity-Ratio" (short AS-Ratio). The ratio is computed by dividing the micro-averaged METEOR score over all regions by an "anatomy-agnostic" METEOR score, calculated when generated sentences of an anatomy are (falsely) paired with reference sentences of all other anatomies of an image. \Eg, if hypothetically there were only the 3 anatomical regions of \emph{right lung} (RL), \emph{left lung} (LL) and \emph{spine} (SP), then the "anatomy-agnostic" score would be calculated by pairing the generated sentences of RL with the reference of LL and SP, the generated sentences of LL with the reference of RL and SP etc., for a given image. Consequently, the AS-Ratio measures how closely aligned the generated sentences are to their corresponding anatomy, with a ratio of 1.0 representing no apparent relation, while higher ratios represent closer alignment.

\noindent\textbf{Selection-based sentence generation.} Quantitatively evaluating this task poses a challenge, since there is no strict ground-truth for sentences generated from manually drawn boxes. As such, we decided to evaluate this task by investigating the impact that randomly varying bounding boxes in relation to ground-truth bounding boxes of anatomical regions have on the METEOR scores (thereby simulating radiologists manually drawing bounding boxes around anatomical regions). Specifically, we vary bounding boxes by either position, aspect ratio, or scale, as illustrated in \cref{fig:bbox-variations}. Intuitively, we want to verify that small deviations from the ground-truth would have a negligible impact on the quality of the generated sentences, while large deviations would have a noticeable negative impact, thereby verifying that selection-based sentence generation is robust to different forms of drawn boxes while still being sensitive to the rough position of the selected region. We conduct several runs sampling from a normal distribution with increasingly higher standard deviations to vary the bounding boxes, thus achieving ever larger deviations from the ground-truth. We refer to \cref{appendix:variation_sampling} for details.

\begin{figure}[t!]
  \centering
   \includegraphics[clip, trim=0cm 23.25cm 12.55cm 0cm, width=1.0\linewidth]{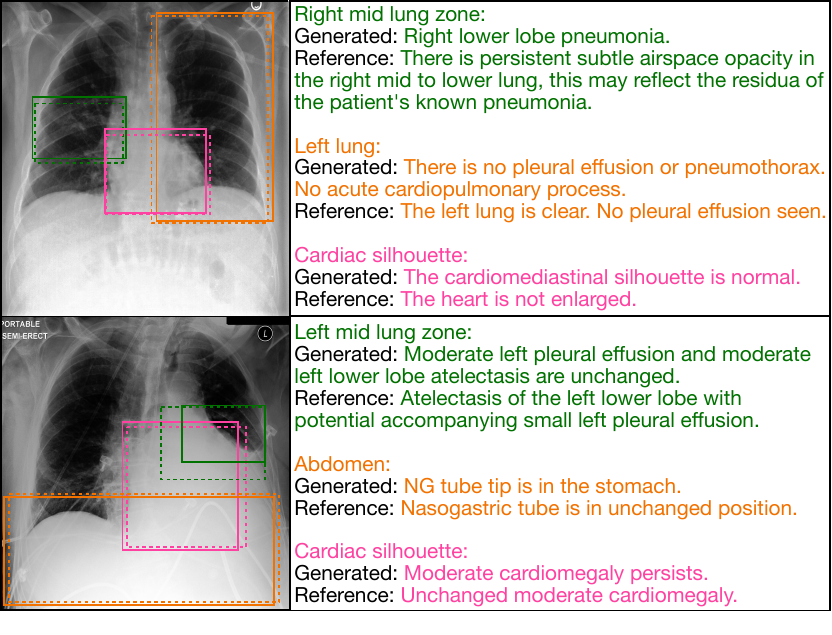}
   \caption{Qualitative results of anatomy-based sentence generation, including predicted (dashed boxes) and ground-truth (solid boxes) anatomical regions. Sentences are color-coded according to their corresponding anatomical regions. We observe that the model generates pertinent, anatomy-related sentences.}
   \label{fig:qualitative-analysis}
\end{figure}

\section{Results and discussion}

\subsection{Radiology report generation}

Our model shows excellent radiology report generation performance (see \cref{tab:NLG-metrics} and \cref{tab:CE-metrics-total}), either being competitive with or outperforming previous models on conventional NLG metrics -- setting a new state-of-the-art (SOTA) on the METEOR metric -- as well as clinically relevant CE metrics -- outperforming all models not directly optimized for these metrics by large margins.

On the BLEU scores, our model is competitive with the latest SOTA models \cite{wang2022inclusive, nicolson2022improving}. However, it is worth noting that \cite{nicolson2022improving} and \cite{wang2022inclusive} applied further processing such as lowercasing to the reference reports, which is well known to significantly increase BLEU scores \cite{post2018call} (see more details in \cref{appendix:report_processing}). In comparison to the best baseline \cite{miura2021improving} without processed reference reports (highlighted by dashed lines), our model improves the BLEU-4 score by $\Delta+$10.5\%. On the METEOR score, which is case-insensitive, our model outperforms the latest SOTA model \cite{nicolson2022improving} by $\Delta+$9.8\% and achieves a $\Delta+$6.3\% increase against the best baseline \cite{you2021aligntransformer}. The score in the ROUGE-L (F1) metric, a summarization evaluation metric, is noticeably lower compared to the baselines. We believe that this can be attributed to the low precision score of the region selection module (see \cref{region_selection_abnormality_classifier_results}), \ie more regions tend to be selected for a generated report than are described in the reference report (in turn causing a decrease in ROUGE-L precision).

For the micro-averaged CE metrics, our model achieves a substantial $+$10.0\% increase ($\Delta+$22.4\%) in F1 score against the best baseline \cite{miura2021improving} (highlighted with dashed lines) not optimized via reinforcement learning (RL). Compared to the best baseline for the example-based CE metrics, our model again shows strong improvements with a $+$6.3\% increase ($\Delta+$16.4\%) in F1 score. Our model is outperformed yet competitive with the baseline \cite{miura2021improving} explicitly optimized on CE metrics via RL, which demonstrates the effectiveness of our divide-and-conquer approach towards generating complete and consistent reports. 

\subsection{Anatomy-based sentence generation}

Our model generates pertinent, anatomy-related sentences, as verified by quantitative (\cref{tab:meteor-region-results}) and qualitative (\cref{fig:qualitative-analysis}) results.

We showcase generated sentences for various selected anatomies of two test set images in \cref{fig:qualitative-analysis}. Predicted bounding boxes closely align with the ground-truth in most cases, which is consistent with the good quantitative results of the object detector (we refer to \cref{appendix:od_results}).

The anatomy-sensitivity observed in the qualitative results is further verified by the Anatomy-Sensitivity-Ratio of 1.938 in \cref{tab:meteor-region-results}, which means that the METEOR score for sentences generated for a given anatomy is almost twice the score for when the model would have generated anatomy-agnostic sentences. It is noteworthy that generated descriptions of abnormal regions tend to reference earlier examinations (hence "\emph{unchanged}" or "\emph{persists}"), which can be explained by the sequential nature of examinations, especially if the progression of a disease is being tracked. Also, we observed that abnormal regions tend to have more diverse reference descriptions than normal regions. This in turn may cause lower scores in conventional NLG metrics for abnormal regions compared to normal regions (see \cref{tab:meteor-region-results}), even though the generated description may be clinically accurate (see "\emph{right mid lung zone}" in the upper image of \cref{fig:qualitative-analysis}).

\begin{table}[t!]
\begin{subtable}{1\linewidth}
\centering
\resizebox{\linewidth}{!}{
\begin{tabular}{lllllll} \hline
Region & RL    & LL    & SP    & MED   & CS    & AB    \\ \hline
METEOR                 & 0.104 & 0.105 & 0.165 & 0.119 & 0.110 & 0.119 \\ \hline                  
\end{tabular}
}
\caption{Six prominent regions: \emph{right lung} (RL), \emph{left lung} (LL), \emph{spine} (SP), \emph{mediastinum} (MED), \emph{cardiac silhouette} (CS) and \emph{abdomen} (AB).}
\label{tab:meteor-region}
\end{subtable}

\vspace{1em}

\begin{subtable}{1\linewidth}
\small
\centering
\begin{tabular}{llll||l} \hline
Region Subset & All & Normal & Abnormal & AS-Ratio \\ \hline
METEOR                 & 0.115   & 0.202 & 0.064 & 1.938 \\ \hline                  
\end{tabular}
\caption{Micro-average over all, normal, and abnormal regions, respectively, and AS-Ratio.}
\label{tab:meteor-region-subset}
\end{subtable}
\caption{Language model results on the anatomy-based sentence generation task. We report (a) the per-anatomy METEOR scores for six prominent regions as well as (b) region subsets and the Anatomy-Sensitivity-Ratio (AS-Ratio). These results verify that our model generates anatomy-specific sentences as the AS-Ratio is two times higher than it would be for anatomy-agnostic sentences.}
\label{tab:meteor-region-results}
\end{table}

\subsection{Selection-based sentence generation}
Our model shows high robustness towards deviations from the ground-truth in aspect ratio and scale while being sensitive to the position (see 
\cref{fig:bbox-variations-plot}). This implies that when a region is specifically targeted for examination by manual bounding box annotation, the radiologist only has to position the bounding box close to the region of interest without worrying too much about capturing the exact aspect ratio and scale of the region. Thus, the integration of selection-based sentence generation as an interactive component into a clinical setting is viable.

\begin{figure}[t!]
  \centering
   \includegraphics[clip, trim=4.75cm 8.8cm 5cm 10.6cm, width=.85\linewidth]{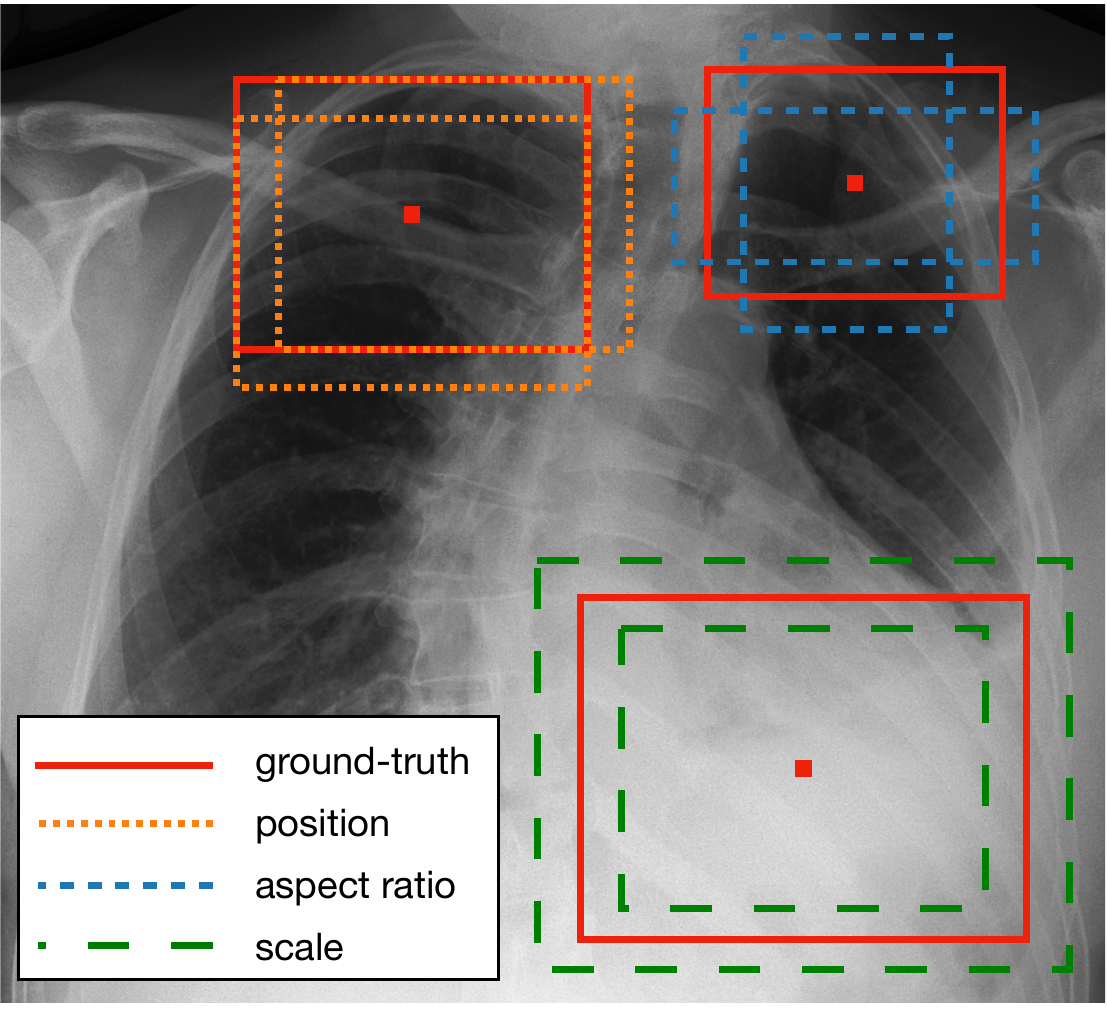}
   \caption{Illustration of the different bounding box variations used to evaluate the model's performance in generating selection-based sentences (\ie, generated for manually drawn bounding boxes). 
   We randomly vary the position, aspect ratio, and scale in order to simulate deviations of manually drawn bounding boxes from ground-truth boxes.}
   \label{fig:bbox-variations}
\end{figure}

\begin{figure}[t!]
  \centering
   \includegraphics[width=1.\linewidth]{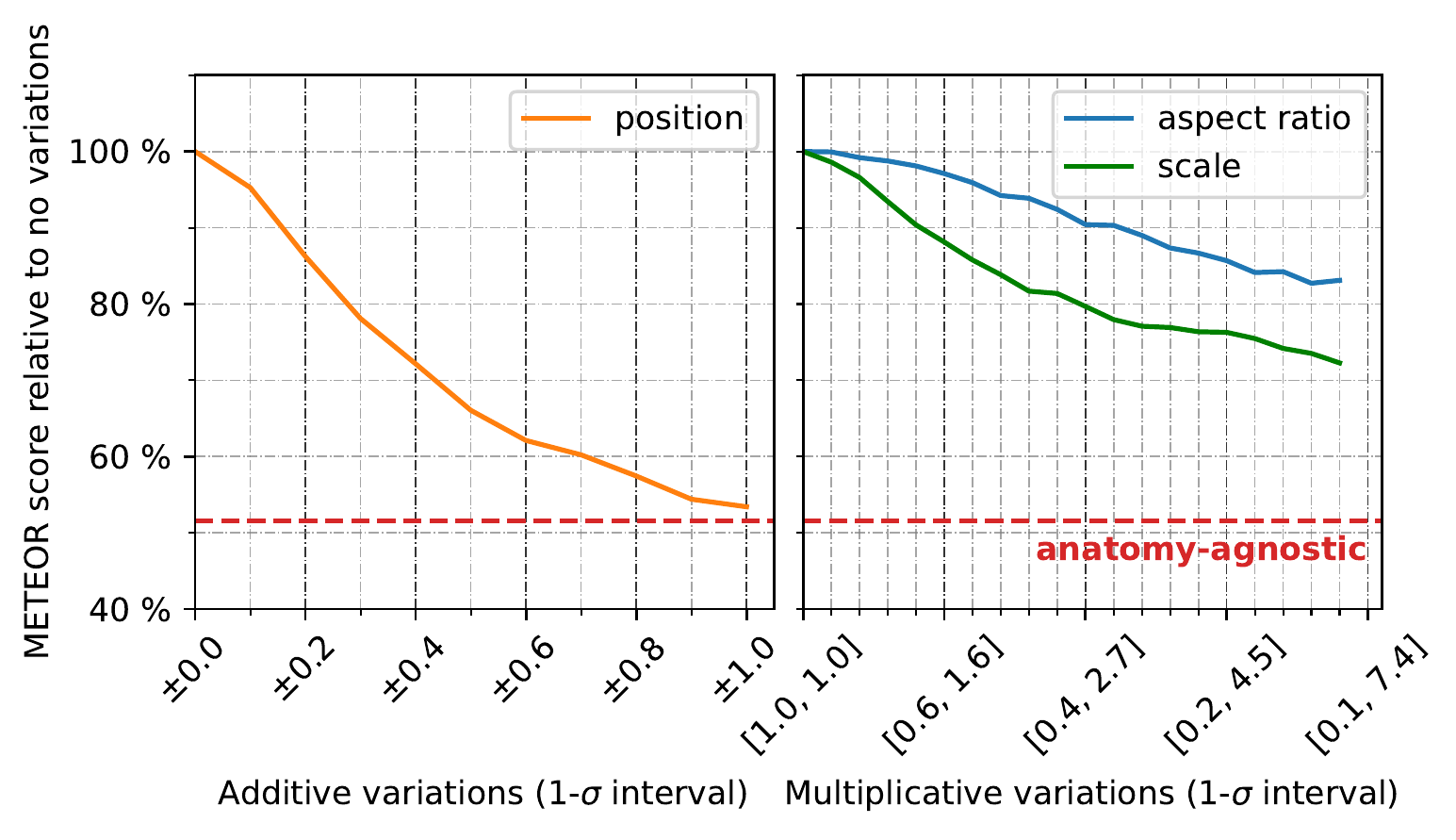}
   \caption{METEOR score of sentences generated from bounding boxes that increasingly deviate from the ground-truth to simulate manually drawn boxes (see \cref{fig:bbox-variations}). For reference, we also show the anatomy-agnostic METEOR score (red) for highlighting the threshold when anatomy-specificity is no longer achieved. Selection-based sentence generation is more sensitive to position and robust towards aspect ratio and scale, demonstrating the viability of this interactive application.}
   \label{fig:bbox-variations-plot}
\end{figure}

\cref{fig:bbox-variations-plot} showcases this effect by plotting the change in METEOR score when manually drawn bounding boxes increasingly deviate from the ground-truth for different variation types (see \cref{fig:bbox-variations}). Since aspect ratio and scale are varied multiplicatively (with 1.0 representing no variation), the scores are plotted against ranges of multiplicative factors, whereas position is varied additively (with 0.0 representing no variation), thus the scores are plotted against increments of additive factors. We observe that the performance declines very slowly when the aspect ratio is varied and somewhat more when the scale is varied, while for position variations the model quickly reaches the score threshold at which generated sentences are considered anatomy-agnostic (see AS-Ratio presented in \cref{tab:meteor-region-results}).

\subsection{Component analysis and ablation study}
We conduct an ablation study on the region selection and abnormality classification modules (\cref{appendix:ablation_study}), verifying their effectiveness and relevance for the overall model performance. In addition, we provide quantitative results and discussion of specific modules (\cref{appendix:od_results,region_selection_abnormality_classifier_results}), and present further qualitative and fine-grained quantitative results of our method (\cref{appendix:additional_results}).

\section{Conclusion}

In this paper, we present a simple yet effective approach to radiology report generation by explicitly focusing on salient anatomical regions through object detection and generating region-specific descriptions. Our method provides a high degree of explainability by visually grounding generated sentences on anatomical regions. Novel interactive capabilities allow radiologists direct involvement in the decision-making process (\eg, by selecting a subset of anatomies or drawing bounding boxes for sentence generation). Such interactivity is crucial for successful integration into a real clinical setting. The experiments verify the effectiveness of our method for generating clinically accurate reports and offering these advantageous interactive capabilities. We hope that our work encourages further research towards region-guidance in radiology report generation.

\noindent\textbf{Ethical Considerations.} While automatic radiology report generation has the potential to substantially improve efficiency in the clinical workflow, erroneous diagnostic evaluations by such systems can have direct, harmful consequences for patients. As the clinical accuracy of automatic diagnostic methods improves, we see an increased risk of overreliance \cite{passi2022overreliance} 
by automation bias (tendency to favor machine-generated decisions while disregarding contrary data \cite{geis2019ethics}). Interestingly, detailed explanations provided by a system may even exacerbate automation bias \cite{schaffer2019can}. However, we believe that our approach of explainability through visual grounding and additional interactive capabilities encourages manual verification and intervention by radiologists, which may mitigate the risk of overreliance.

\noindent\textbf{Limitations.} While technically simple, our method requires strong supervision, currently only provided by the Chest ImaGenome dataset, and is thus hard to generalize to other report generation tasks. Adapting our approach for limited supervision is a promising future research direction. Secondly, our method only considers chest X-rays in isolation, whereas chest X-rays are usually diagnostically evaluated in the context of previous examinations. 
The Chest ImaGenome dataset contains localized comparison relations for anatomical regions across sequential exams, and incorporating this information has the potential to improve the clinical accuracy of generated reports. Finally, the reference reports of the MIMIC-CXR dataset contain sentences describing objects that cannot be attributed to an anatomical region (\eg surgical clips) and are thus not covered by our method.  
Future work may thus consider a hybrid system utilizing region- and image-level visual features to generate both region- and such image-level sentences.

{\small
\bibliographystyle{ieee_fullname}
\bibliography{ms}
}
\clearpage
\appendix
\begin{table*}[htb!]
\centering
\resizebox{\textwidth}{!}{
\begin{tabular}{c|cccc|ccccc}
\hline
Dataset                    & \makecell{Object \\ detector}         & \makecell{Abnormality \\ classification} & \makecell{Region \\ selection} & \makecell{Language \\ model} & BLEU-4 & METEOR & P\textsubscript{mic-5} & R\textsubscript{mic-5} & F\textsubscript{1, mic-5}\\ \hline
\multirow{4}{*}{MIMIC-CXR} & \cmark           &   &   & \cmark  & 0.104 &  0.135  & 0.578  & 0.359  & 0.443    \\
                           & \cmark & \cmark  &    &  \cmark  & 0.107 & 0.138  & 0.550  & 0.461  & 0.501    \\ 
                           & \cmark &  & \cmark  & \cmark  & 0.114 & 0.161  & 0.498  & 0.551  & 0.523    \\ \cline{2-10}
                           & \cellcolor{Gray}\cmark          & \cellcolor{Gray}\cmark & \cellcolor{Gray}\cmark      & \cellcolor{Gray}\cmark      & \cellcolor{Gray}\textbf{0.126}  &    \cellcolor{Gray}\textbf{0.168}   & \cellcolor{Gray}0.491   & \cellcolor{Gray}\textbf{0.617}       & \cellcolor{Gray}\textbf{0.547}
                           \\ \hline
\end{tabular}}
\caption{Ablation study on the abnormality classification and region selection modules. The performance is evaluated on two natural language generation metrics (BLEU-4 and METEOR) and clinical efficacy metrics micro averaged over five observations. Each module contributes to an increased performance (especially in recall) of the RGRG model.}
\label{tab:ablation-study}
\end{table*}

\section{Detailed analysis}
\subsection{Ablation study}\label{appendix:ablation_study}

\cref{tab:ablation-study} shows the results of an ablation study on the region selection and abnormality classification modules. For full report generation, our method at minimum requires an object detector to extract region visual features and a language model to generate region-specific sentences, thus these two modules together form the base model. We investigate the effects of incorporating the abnormality classification and region selection module, respectively, into this base model by evaluating on BLEU-4, METEOR, and clinical efficacy (CE) metrics micro-averaged over five observations.

We observe that adding the abnormality classification module has a negligible effect on conventional natural language generation (NLG) metrics of BLEU-4 and METEOR, whilst substantially improving CE recall by $+$10.2\% ($\Delta+$28.4\%) at the slight expense of CE precision. This showcases that 1) conventional NLG metrics are ill-suited for evaluating the clinical accuracy of generated reports \cite{boag2020baselines, liu2019clinically, pino2020inspecting} and 2) the abnormality classification module effectively encodes abnormality information in the region visual features, as evidenced by the substantial increase in recall.

Incorporating the region selection module substantially boosts the performance of the base model across all metrics, likely due to the changed approach in training the language model once the region selection module is introduced to the base model. In the base model, the language model is trained with all reference sentences (\ie, empty and non-empty) of all 29 regions per image, as the generated sentences of all 29 regions are concatenated to form the final report. Since there are 2.2 times more empty reference sentences than non-empty reference sentences (see weighted binary cross-entropy loss of the region selection module in \cref{training}), the language model learns to often generate empty sentences for regions. Thus intuitively, the language model in the base model is not only tasked with generating region-specific sentences, but also with "deciding" which regions require non-empty sentences. In addition, the generated reports of the base model are shorter than those of the base model + region selection module. This is because even though the language model in the base model generates sentences for all 29 regions (which are concatenated to form the final report), a lot of these generated sentences will be empty. We verify this by calculating the average number of tokens (using a Spacy tokenizer) in a generated report by the base model vs. base model + region selection module. While a generated report by the base model contains on average 39 tokens, incorporating the region selection module increases this to 52 tokens. Thus, the base model may not be generating sufficiently long reports containing region-specific sentences that accurately describe abnormalities, which may be reflected in the low CE recall score.

\begin{table*}[t!]
\centering
\begin{tabular}{lllllll|l||p{4cm}} \hline
Region & RL    & LL    & SP    & MED   & CS    & AB    & \textbf{Average} & Avg. num. detected regions \\ \hline
IoU                 & 0.925 & 0.920 & 0.950 & 0.870 & 0.837 & 0.913 & 0.887   & \hfil28.792 \\ \hline                  
\end{tabular}
\caption{Object detector results micro averaged over all anatomical regions as well as 6 prominent regions: \emph{right lung} (RL), \emph{left lung} (LL), \emph{spine} (SP), \emph{mediastinum} (MED), \emph{cardiac silhouette} (CS) and \emph{abdomen} (AB). Almost all 29 anatomical regions are detected per image with adequate IoU scores.}
\label{tab:object-detector-results}
\end{table*}

When the region selection module is incorporated into the base model, the language model is trained exclusively on region visual features with corresponding non-empty reference sentences. This removes the implicit task of "deciding" which regions need sentences, potentially allowing the language model more capacity to generate better region-specific sentences. This could explain the $\Delta+$9.6\% increase in the BLEU-4 score and $\Delta+$19.3\% increase in the METEOR score. Additionally, we can see that compared to the base model, the CE recall improves significantly by $+$19.2\% ($\Delta+$53.5\%), which is likely due to the increased capacity in generating better region-specific sentences, thus more abnormalities are correctly described in the final reports. However, we also observe a noticeable decrease in CE precision score compared to the baseline. This may be attributed to the low precision score of the region selection module w.r.t. normal regions (see \cref{region_selection_abnormality_classifier_results}), leading to more normal regions being described in generated reports and thus increasing the likelihood of false positives (\ie, normal regions being described as abnormal).

Finally, by combining the abnormality classification and region selection modules in the RGRG model (outlined in gray), we again see an increase in CE recall, verifying the effectiveness and relevance of both modules for the overall model performance.

\subsection{Object detector results}\label{appendix:od_results}

We evaluate the object detector via the Intersection over Union (IoU) metric, which we calculate as the sum of the intersection areas divided by the sum of union areas. We use the IoU metric instead of the (in object detection) more commonly used mean Average Precision (mAP) metric, since each anatomical region typically appears exactly once in an image, and never more than once. We report the micro average IoU score over all regions as well as for 6 prominent regions. Additionally, we report the average number of detected regions per image.

The IoU scores in \cref{tab:object-detector-results} demonstrate that anatomical regions are detected adequately, with almost all 29 regions being detected per image with an average IoU score of 0.887. We noticed that the ground-truth bounding boxes in the Chest ImaGenome dataset, which were automatically extracted by a bounding box pipeline, do not always precisely overlap with the real regions, 
which likely negatively impacted the IoU scores. However, since ultimately the goal is to generate consistent anatomy-related sentences (and not perfect object detection), we believe that imperfect object detection is acceptable.

\subsection{Region selection and abnormality classification results}\label{region_selection_abnormality_classifier_results}
\begin{table}[t!]
\centering
\setlength{\tabcolsep}{0.5em}
\begin{tabular}{lllll}
\hline
Module & Regions    & P     & R     & F\textsubscript{1}    \\ \hline
\multirow{3}{*}{Region Selection} &
All       & 0.594 & 0.904 & 0.717 \\
& Normal    & 0.459 & 0.903 & 0.608 \\
& Abnormal  & 1.0   & 0.906 & 0.951 \\ \hline
Abnorm.\ Classifier& All  & 0.354 &  0.911 & 0.510 \\ \hline
\end{tabular}
\caption{Results of the region selection and abnormality classification modules. Salient anatomical regions are selected for the final report with high recall for both normal and abnormal regions, at the expense of precision for normal regions. Anatomical regions are classified as abnormal with high recall but decreased precision.}
\label{tab:region-selection-abnormality-classifier-results}
\end{table}

We evaluate the binary classifiers of the two modules on precision, recall, and F1 score. For the region selection module, a region is deemed positive if it has a corresponding reference sentence, and for the abnormality classification module, a region is positive if it is abnormal as per ground-truth. For region selection, we additionally report the scores for the subsets of normal and abnormal regions.

\cref{tab:region-selection-abnormality-classifier-results} showcases the results. We observe that recall is high for both normal and abnormal regions for region selection, thus regions that are described in the reference report are also usually selected for the generated report. However, precision is low for normal regions, meaning usually more normal regions are selected for the generated report than are described in the reference report. As mentioned in the main paper, this can explain the low score for the ROUGE-L (F1) metric, since the generated report thus contains more information than the reference report, which in turn causes a lower ROUGE-L precision score. However, the decision to describe normal regions (\eg, \emph{"There is no pleural effusion or pneumothorax."}) lies with the radiologist and is arbitrary, since pathology-free regions are not required to be mentioned in a report. Thus, we believe that this rather subjective decision cannot be learned by a model and a low precision score for normal regions is expected.

Precision is 1.0 for abnormal regions since by default abnormal regions are always included in reference reports. Hence, there cannot be any false positives for the abnormal region subset. Consequently, the recall score for the normal and abnormal region subsets cannot be directly compared.

For the abnormality classifier, we observe high recall and low precision. Thus, abnormal regions are usually detected correctly while normal regions are sometimes misclassified.

\FloatBarrier
\clearpage

\section{Detailed results}\label{appendix:additional_results}
\subsection{Qualitative anatomy-based sentence generation results}
\begin{figure}[ht!]
  \centering
   \includegraphics[clip, width=.5\textwidth, trim=4.65cm 6.5cm 4.8cm 0.0cm]
   {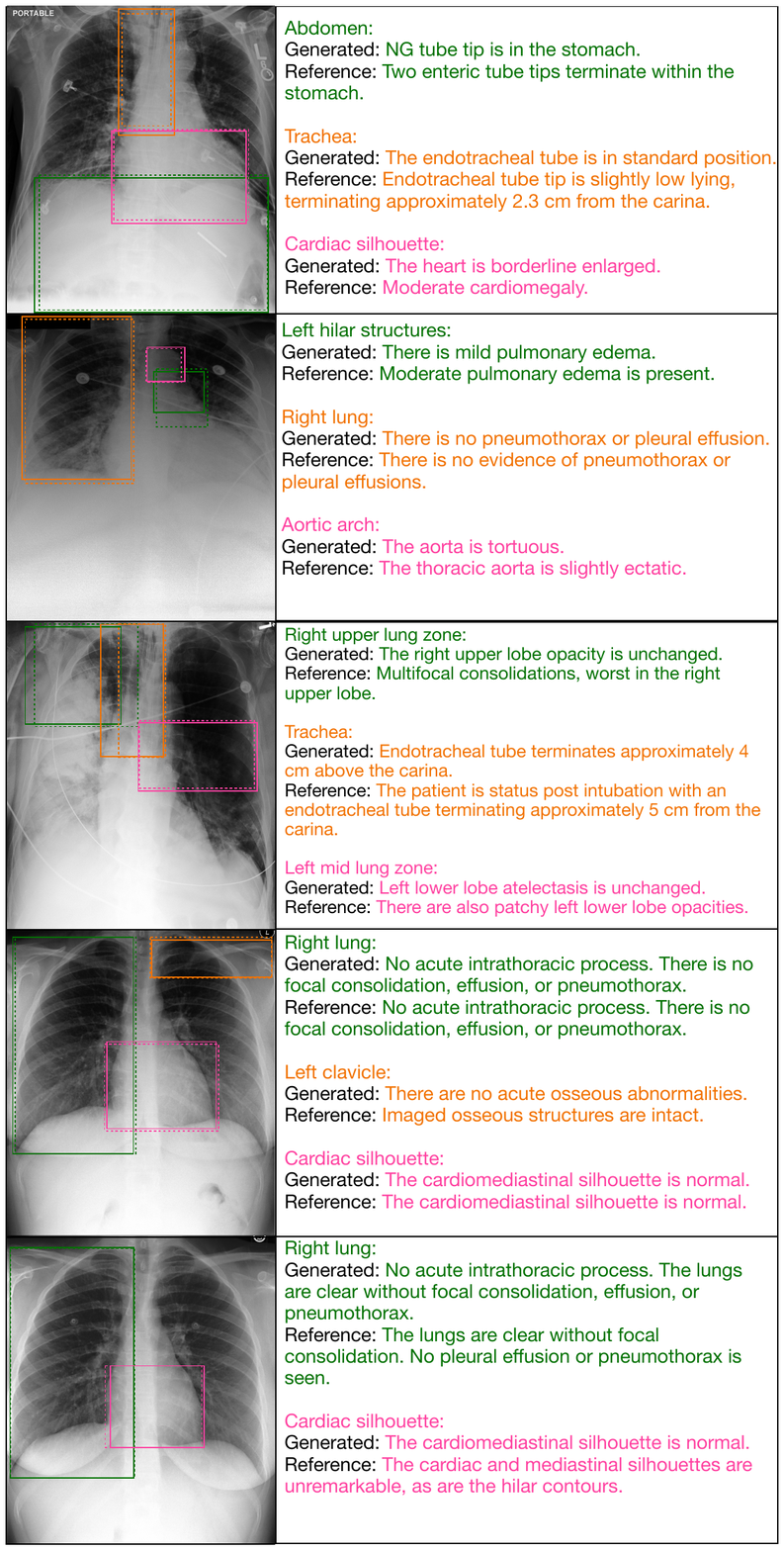}
   \caption{Anatomy-based sentence generation for 5 test set images. The upper three images depict abnormal cases, while the lower two depict normal cases. We show predicted (dashed boxes) and ground-truth (solid boxes) anatomical regions and color sentences accordingly.}
   \label{fig:qualitative_analysis_anatomy_based_generation}
\end{figure}

\subsection{Qualitative full report generation results}
\begin{figure*}[p]
  \centering
   \includegraphics[clip, trim=0cm 11.3cm 0cm 0cm, width=1.0\textwidth]{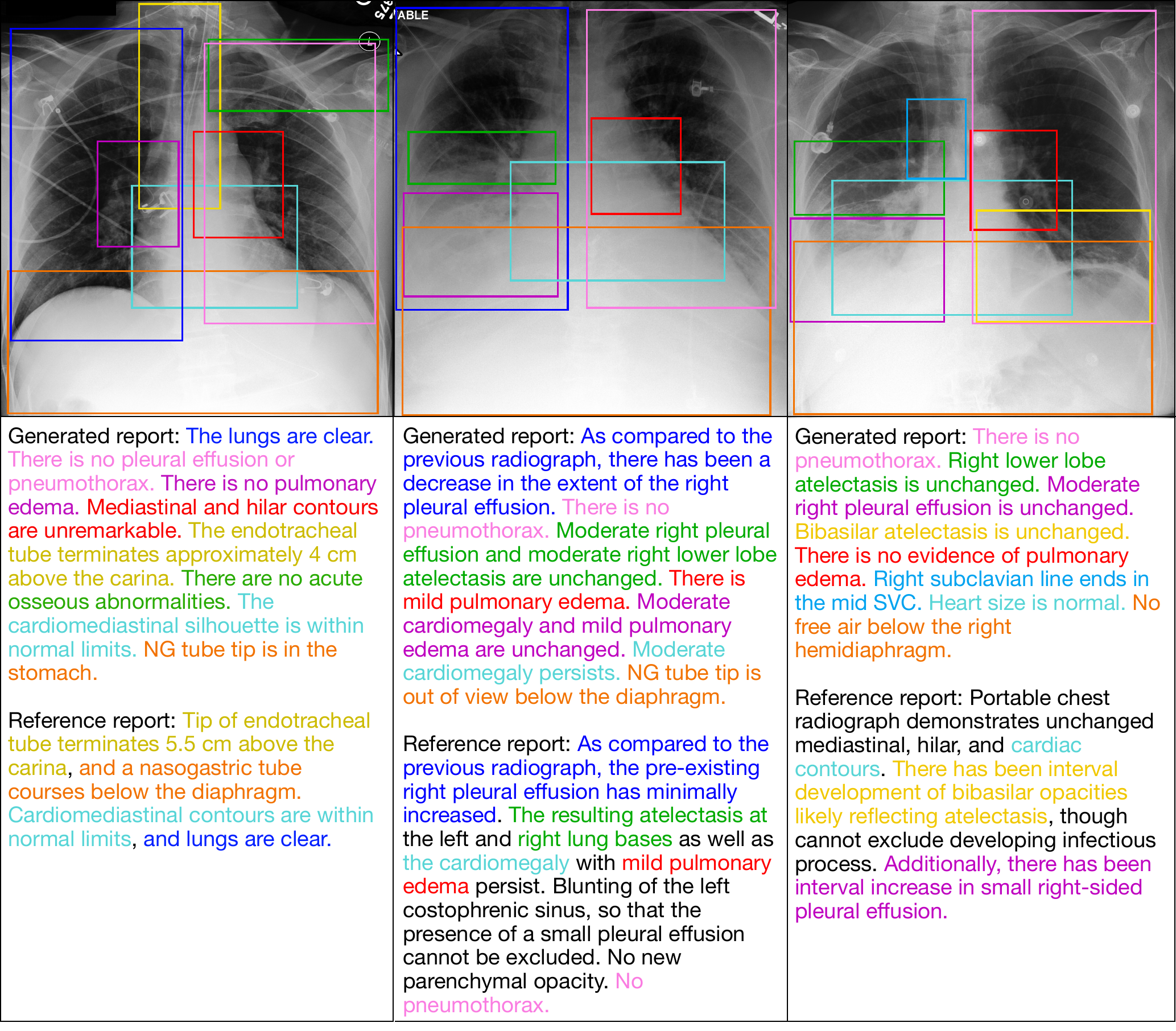}
   \caption{Full report generation for 3 test set images. Detected anatomical regions (solid boxes), corresponding generated sentences, and semantically matching reference sentences are colored the same. The generated reports mostly capture the information contained in the reference reports, as reflected by the matching colors. The left image shows a healthy chest X-ray image
devoid of any pathologies, while the other two images depict abnormalities.}
   \label{fig:qualitative_analysis_report}
\end{figure*}

\cref{fig:qualitative_analysis_report} showcases generated full reports for three test set images. 
The left image shows a healthy chest X-ray image devoid of any pathologies. Based on the matching colors between generated and reference reports, we can see that all information contained in the reference report is also included in the generated report. In particular, the generated report correctly describes the placement of the endotracheal tube (colored in yellow), although with a slightly wrong numerical value. Also, the nasogastric tube is correctly mentioned in the generated report. 
The generated report describes, clinically correctly, four additional negative observations (i.e.\ non-present pathologies), which are however not mentioned in the reference report. As discussed in \cref{region_selection_abnormality_classifier_results}, the region selection module has a low precision score for normal regions, since the decision to mention normal regions in a report is arbitrary and cannot be effectively learned. Thus, typically more regions are described in generated reports than in the corresponding reference reports, which in turn lowers the ROUGE-L score. However, the additionally described observations in the generated report of the left image are all clinically accurate, thus we believe that the region selection module selecting more regions than are described in the reference report is not detrimental to the quality of the generated reports.

In the middle image, we can see that the generated report correctly describes the pleural effusion in the right lung (colored in blue). However, while the generated sentence erroneously specifies a decrease in the effusion in comparison to a previous radiograph, the reference describes an increase. As mentioned in the limitations section of the main paper, our method considers each chest X-ray in isolation, and, as illustrated by this case, cannot correctly generate sentences that depend on previous radiographs. 
Thus, incorporating the information of localized comparison relations for anatomical regions between sequential exams into our method may be required to improve the generation of such sentences. In addition, there are some duplicate mentions of observations in the generated report, but since they are consistent with each other and clinically accurate, this is acceptable from a clinical point of view. However, the generated report misses a potential small pleural effusion on the left side (\emph{"presence of a small pleural effusion cannot be excluded"}), illustrating the need for interactiveness and transparency during report generation, which is simplified by our method.

The right image shows another chest X-ray with pathological findings, which were mainly captured by the generated report. However, we can see that all sentences in the reference report refer to previous radiographs, highlighting again the importance of incorporating sequential information in the method.

\FloatBarrier

\onecolumn
\subsection{Qualitative selection-based sentence generation results}
\begin{figure*}[ht!]
  \centering
   \includegraphics[clip, trim=0cm 18.75cm 1.3cm 0cm, width=1.0\textwidth]{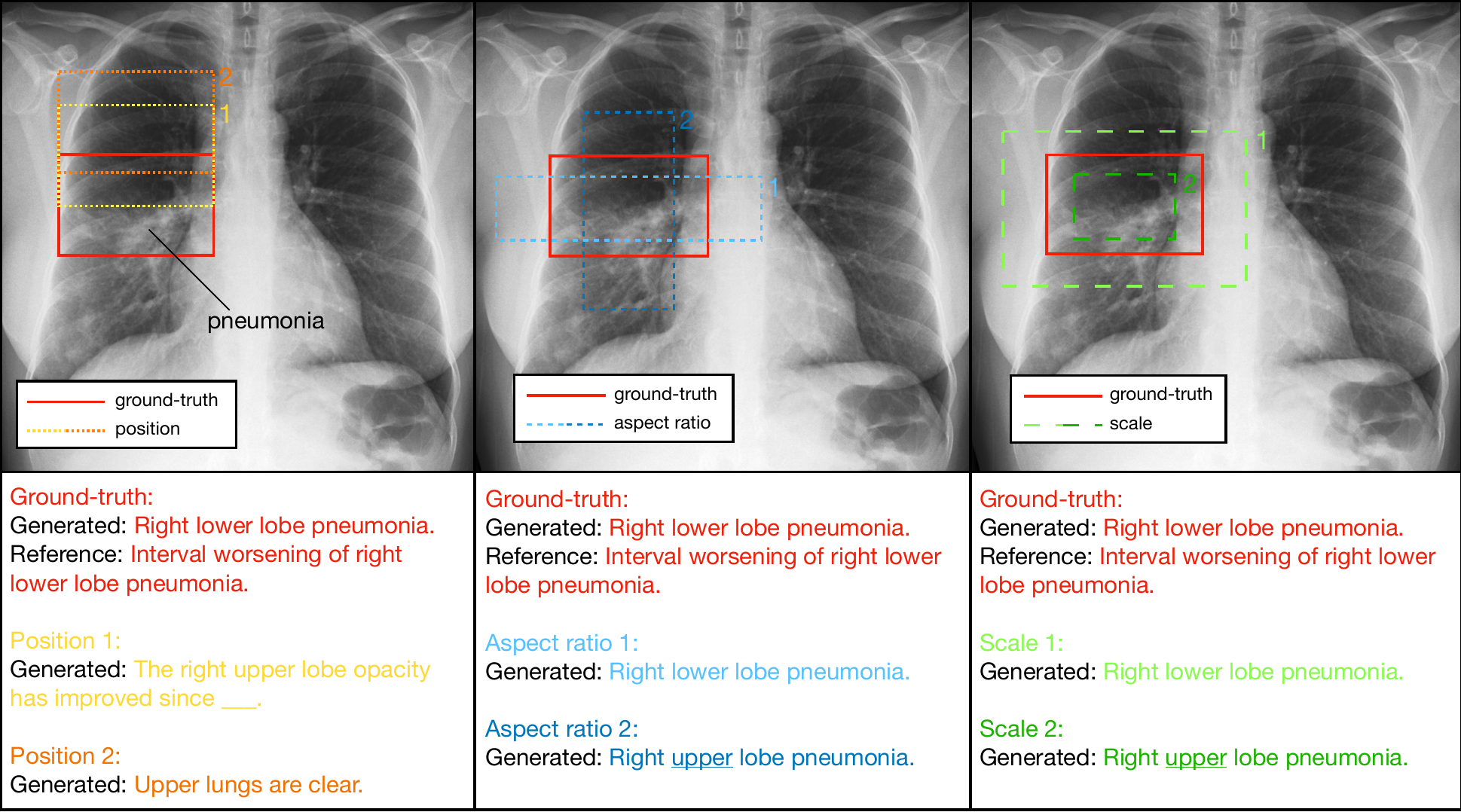}
   \caption{Visualizing selection-based sentence generation for a test set image with pneumonia pathology. The solid red bounding box indicates the ground-truth anatomical region containing the pathology. Various dashed and colored bounding boxes represent radiologist-drawn bounding boxes, deviating from the ground-truth in terms of position, aspect ratio, or scale. The generated sentences demonstrate heightened sensitivity to bounding box position, while maintaining robustness towards variations in aspect ratio and scale.}
   \label{fig:qualitative_analysis_selection_based}
\end{figure*}
\FloatBarrier
\begin{multicols}{2}
\cref{fig:qualitative_analysis_selection_based} showcases the sensitivity of selection-based sentence generation to the position, aspect ratio, and scale of manually drawn bounding boxes within a test set image featuring pneumonia pathology.

The left image in the figure demonstrates variations in the position of the manually drawn bounding boxes. It is evident that the position is crucial, as the generated sentence for position 1 (slightly above the pathology) already misses the pathology and only describes an upper lobe opacity (which we believe is accurate). However, the generated sentence for position 2, which is even higher, completely misses the pathology and states that the upper lungs are clear (which we again believe to be accurate). Consequently, radiologists must be cautious to accurately position the bounding box to ensure correct pathology detection.

The middle image in the figure displays variations in the aspect ratio of the bounding boxes. For aspect ratios 1 and 2, both generated sentences correctly identify pneumonia. However, the sentence for aspect ratio 2 erroneously indicates that the pneumonia is located in the right upper lobe, rather than the lower lobe. This mistake is understandable, as the bounding box lacks sufficient surrounding information to accurately determine the relative position within the lung (\ie, upper or lower lobe).

The right image in the figure showcases variations in scale. The generated sentences exhibit robustness, as both scale variations correctly identify pneumonia. However, similar to the aspect ratio case, the sentence for scale 2 inaccurately describes an upper lobe pneumonia. Again, this error can be attributed to the insufficient surrounding information in the small-scaled bounding box.

In conclusion, selection-based sentence generation introduces additional flexibility into the clinical workflow by allowing radiologists to draw bounding boxes around areas of interest anywhere in the image. The primary caveat is the importance of correct positioning for the bounding box, which, if possible, should contain enough surrounding information to enable the model to generate accurate sentences.
\end{multicols}
\FloatBarrier
\onecolumn
\subsection{Detailed clinical efficacy metrics results}
{
\captionsetup{type=table}
\centering
\resizebox{.75\textwidth}{!}{
\begin{tabular}{c|l|cccc}
\hline
\multirow{2}{*}{Dataset}    & \multicolumn{1}{c|}{\multirow{2}{*}{Observation}} & \multicolumn{4}{c}{{RGRG}}                                                                                \\ 
\cline{3-6} 
                            & \multicolumn{1}{c|}{}                             & \multicolumn{1}{c|}{P}     & \multicolumn{1}{c|}{R}     & \multicolumn{1}{c|}{F\textsubscript{1}}             & acc.      \\ \hline
\multirow{15}{*}{MIMIC-CXR} & Micro Average                                     & \multicolumn{1}{c|}{0.524} & \multicolumn{1}{c|}{0.474} & \multicolumn{1}{c|}{0.498}          & 0.849          \\ \cline{2-6}
                            & Atelectasis                                       & \multicolumn{1}{c|}{0.402} & \multicolumn{1}{c|}{0.853} & \multicolumn{1}{c|}{0.546} & 0.602     \\
                            & Cardiomegaly                                      & \multicolumn{1}{c|}{0.577} & \multicolumn{1}{c|}{0.679} & \multicolumn{1}{c|}{0.624} & 0.770    \\
                            & Consolidation                                     & \multicolumn{1}{c|}{0.132} & \multicolumn{1}{c|}{0.055} & \multicolumn{1}{c|}{0.078}          & 0.919        \\ 
                            & Edema                                             & \multicolumn{1}{c|}{0.504} & \multicolumn{1}{c|}{0.524} & \multicolumn{1}{c|}{0.514}          & 0.859      \\ 
                            & Pleural Effusion                                  & \multicolumn{1}{c|}{0.700} & \multicolumn{1}{c|}{0.467} & \multicolumn{1}{c|}{0.560}          & 0.826\\ \cline{2-6} 
                            & Enlarged Cardiomediastinum                        & \multicolumn{1}{c|}{0.360} & \multicolumn{1}{c|}{0.001} & \multicolumn{1}{c|}{0.003}          & 0.811   \\
                            & \MySmallestIndent\cellcolor{Gray} Fracture                                          & \multicolumn{1}{c|}{\cellcolor{Gray}0.0}   & \multicolumn{1}{c|}{\cellcolor{Gray}0.0}   & \multicolumn{1}{c|}{\cellcolor{Gray}0.0}    &     \cellcolor{Gray}0.0    \\
                            & Lung Lesion                                       & \multicolumn{1}{c|}{0.217} & \multicolumn{1}{c|}{0.004} & \multicolumn{1}{c|}{0.007}          & 0.957 \\
                            & Lung Opacity                                      & \multicolumn{1}{c|}{0.517} & \multicolumn{1}{c|}{0.181} & \multicolumn{1}{c|}{0.268}          & 0.730  \\
                            & No Finding                                        & \multicolumn{1}{c|}{0.554} & \multicolumn{1}{c|}{0.735} & \multicolumn{1}{c|}{0.632} & 0.805     \\
                            & Pleural Other                                     & \multicolumn{1}{c|}{0.200} & \multicolumn{1}{c|}{0.001} & \multicolumn{1}{c|}{0.002} & 0.975       \\ 
                            & Pneumonia                                         & \multicolumn{1}{c|}{0.240} & \multicolumn{1}{c|}{0.122} & \multicolumn{1}{c|}{0.162}          & 0.880        \\
                            & Pneumothorax                                      & \multicolumn{1}{c|}{0.189} & \multicolumn{1}{c|}{0.138} & \multicolumn{1}{c|}{0.159}          & 0.950   \\ 
                            & Support Devices                                   & \multicolumn{1}{c|}{0.732} & \multicolumn{1}{c|}{0.687} & \multicolumn{1}{c|}{0.709} & 0.838 \\ \hline
\end{tabular}
}
\caption{Detailed results for the clinical efficacy (CE) metrics (see \cref{clinical-efficacy-metrics} for details) for each observation as well as micro averaged over all 14 observations. The first five observations listed from the top are those used in calculating the P\textsubscript{mic-5}, R\textsubscript{mic-5}, and F\textsubscript{1, mic-5} scores in Tab. 2 of the main paper. The observation of fracture has a score of 0.0 (outlined in gray), since there are no sentences describing fractures in the Chest ImaGenome dataset. Thus, as mentioned in the limitations section of the main paper, a hybrid model that uses image-level features and sentences describing observations such as fractures from the MIMIC-CXR dataset may be required to further improve clinical accuracy.
}
\label{tab:detailed-ce-results}
}

\subsection{Detailed anatomy-level results}
{
\captionsetup{type=table}
\centering
\resizebox{\textwidth}{!}{
\begin{tabular}{c||l|c|c||l|c|c}
\hline
Dataset                           & Anatomical Region                      & METEOR & IoU    & Anatomical Region & METEOR & IoU   \\ \hline
\multirow{15}{*}{Chest ImaGenome} & Abdomen                                & 0.119  & 0.913  & Right Apical Zone        & 0.157  & 0.863 \\
                                  & Aortic Arch                            & 0.127  & 0.759  & Right Atrium             & 0.237  & 0.755 \\
                                  & Cardiac Silhouette                     & 0.110  & 0.837  & Right Clavicle           & 0.290  & 0.849 \\
                                  & Carina                                 & 0.229  & 0.542  & Right Costophrenic Angle & 0.264  & 0.819 \\
                                  & Cavoatrial Junction                    & 0.171  & 0.616  & Right Hemidiaphragm      & 0.147  & 0.826 \\
                                  & Left Apical Zone                       & 0.157  & 0.873  & Right Hilar Structures   & 0.104  & 0.882 \\
                                  & Left Clavicle                          & 0.294  & 0.841  & Right Lower Lung Zone    & 0.051  & 0.897 \\
                                  & Left Costophrenic Angle                & 0.270  & 0.858  & Right Lung               & 0.104  & 0.925 \\
                                  & Left Hemidiaphragm                     & 0.074  & 0.796  & Right Mid Lung Zone      & 0.083  & 0.893 \\
                                  & Left Hilar Structures                  & 0.108  & 0.875  & Right Upper Lung Zone    & 0.066  & 0.920 \\
                                  & Left Lower Lung Zone                   & 0.054  & 0.881  & Spine                    & 0.165  & 0.950 \\
                                  & Left Lung                              & 0.105  & 0.920  & SVC                      & 0.162  & 0.790 \\
                                  & Left Mid Lung Zone                     & 0.089  & 0.894  & Trachea                  & 0.144  & 0.857 \\
                                  & Left Upper Lung Zone                   & 0.049  & 0.922  & Upper Mediastinum        & 0.162  & 0.881 \\
                                  & Mediastinum                            & 0.119  & 0.870                    \\\hline                   
\end{tabular}
}
\caption{Detailed results of the anatomy-based sentence generation (evaluated using METEOR) and object detection (evaluated using the IoU score) for each of the 29 anatomical regions.}
\label{tab:detailed-anatomy-level-results}
}

\twocolumn
\section{Method}\label{appendix:method}

\subsection{Module details}

\noindent\textbf{Object detector.} Readers familiar with the Faster R-CNN \cite{ren2015faster} architecture may wonder why our method does not use RoI feature vectors (which are extracted from the RoI feature maps through fully connected layers in Faster R-CNN) directly as our region visual features, since instead we apply 2D average pooling and a linear transformation to the RoI feature maps to extract the region visual features. We found that taking the RoI feature vectors directly as the region visual features hurt the object detector's performance, which we suspect is due to the coupling of features between the object detector’s subsequent classifier and regressor and the report generation model’s subsequent modules.

\noindent\textbf{Language model.} For the language model, we use the GPT-2 implementation from the huggingface library (transformers 4.19.2) \cite{wolf2020transformers} with the following checkpoint \cite{papanikolaou2020dare}: \url{https://huggingface.co/healx/gpt-2-pubmed-medium}.

\subsection{Training}
\label{training}

For the overall training loss (Eq.\ 3 of the main paper), we specified that $\mathcal{L}_{\text{select}}$ and $\mathcal{L}_{\text{abnormal}}$ are weighted binary cross-entropy losses for the region selection and abnormality classification modules. Based on statistics computed on the training dataset, these weights for the positive examples are set to 2.2 for $\mathcal{L}_{\text{select}}$ and 6.0 for $\mathcal{L}_{\text{abnormal}}$ to account for class imbalances between regions with/without sentences and that are abnormal/normal, respectively.

As mentioned in the main paper, the model is trained in three stages:

\begin{enumerate}
    \item Object detector
    \item Object detector + region selection module \\ + abnormality classification module
    \item Full model end-to-end
\end{enumerate}

During all three stages, we train on a single NVIDIA A40 with PyTorch 1.12.1 in native mixed precision. The total training took about 45 hours and up to 48 GB of GPU memory was required. We refer to the code for more specifications of dependencies and versions. We use the AdamW \cite{loshchilov2018decoupled} optimizer with a weight decay of 1e-2,
reduce the learning rate by a factor of 0.5 if the total validation loss has plateaued or decreased (compared to the best epoch), and apply early stopping.
In the first training stage, we use a batch size of 16, an initial learning rate of 1e-3, and train for 6 epochs.
In the second training stage, we use a batch size of 16, an initial learning rate of 5e-4, and train for 9 epochs.
In the third training stage, we use a batch size of 2, an initial learning rate of 5e-5, and train for 2 epochs.
All batch sizes are (gradient) accumulated to 64.

\subsection{Inference}

\noindent\textbf{Sentence generation.} We employ beam search with a width of 4 for sentence generation and use a BERTScore \cite{zhang2019bertscore} threshold of 0.9 (based on best validation set performance) to remove similar generated sentences in radiology report generation. The high BERTScore value ensures robust duplicate removal, as only highly similar sentences are deduplicated, minimizing the risk of eliminating relevant information. For BERTScore, we use the uncased base version of DistilBERT \cite{sanh2019distilbert} (\emph{distilbert-base-uncased}).

\section{Experimental Setup}\label{appendix:exp_setup}

\subsection{Dataset and pre-processing}

We use the recently released Chest ImaGenome v1.0.0 \cite{wu2021chest, wu2021chestphysio, PhysioNet} dataset for training and evaluation of our proposed model. The MIMIC-CXR \cite{johnson2019mimic, johnson2019mimicphysio} dataset, from which the Chest ImaGenome dataset is automatically constructed, consists of 377,110 chest X-ray images corresponding to 227,835 free-text radiology reports. The Chest ImaGenome contains automatically constructed scene graphs for 242,072 of those MIMIC-CXR images. For the images themselves, we use the MIMIC-CXR-JPG v2.0.0 \cite{johnson2019mimicjpg, johnson2019mimicjpgphysio} dataset, which is fully derived from MIMIC-CXR and conveniently offers the images in JPG format.

The following image data augmentations are applied with 50\% probability (each) during training:

\begin{itemize}
    \item Color jitter of 20\% brightness and contrast (saturation and hue jittering are not used as chest X-rays are single-channel greyscale images)
    \item Gaussian noise of zero mean and variance in the range [10, 50]
    \item Affine transformation with translation up to $\pm$2\% of the image height/width and rotation up to $\pm$2°
\end{itemize}

For the sentences of the Chest ImaGenome dataset, we always remove redundant whitespaces (as mentioned in the main paper). In some cases, we noticed that sentences assigned to regions contained superfluous, introductory phrases (such as \emph{“UPRIGHT PORTABLE AP CHEST RADIOGRAPH:”}), which do not contain any relevant information. We assumed that these phrases were erroneously extracted by the Chest ImaGenome dataset from the MIMIC-CXR radiology reports and assigned to regions, thus they are also removed.

\subsection{Reference reports and processing}\label{appendix:report_processing}

As described in the main paper, we use the \emph{findings} section of MIMIC-CXR radiology reports as our reference reports. To extract these sections, we use a text extraction tool provided by the MIMIC-CXR dataset authors: \url{https://github.com/MIT-LCP/mimic-cxr/tree/master/txt}.

We emphasize that we do not apply any further processing to these extracted reports. In contrast, some papers, such as the two papers \cite{nicolson2022improving, wang2022inclusive} from 2022 in Tab. 1 of the main paper, most likely applied additional processing to these extracted reports, including lowercasing all words. While \cite{nicolson2022improving} details the applied processing, \cite{wang2022inclusive} does not provide this information, and no code is available for verification. However, their qualitative analysis showcases lowercased reference reports, leading us to believe that they did employ lowercasing.

Lowercasing can significantly impact natural language generation (NLG) scores, particularly BLEU scores \cite{post2018call}. We discovered that when lowercasing reference reports, our method produces these BLEU scores: \emph{BLEU-1: 0.400, BLEU-2: 0.266, BLEU-3: 0.187, BLEU-4: 0.135 ($\Delta$+8.9\% against best baseline)}. METEOR and CE scores remain unchanged, as lowercasing does not affect them.

We believe that this highlights another reason why NLG metrics are ill-suited for evaluating radiology reports, as scores heavily depend on the specific processing applied to reference reports (since NLG metrics count matching n-grams). In contrast, CE-metrics are processing-invariant, as they compare disease presence status between reference and generated reports, independent of sentence structure or casing. Thus, CE metrics allow for a fairer comparison between methods while also capturing the diagnostic accuracy of generated reports. Consequently, we encourage future radiology report generation research to place greater emphasis on CE metrics when evaluating generated reports.

\subsection{Clinical efficacy metrics}
\label{clinical-efficacy-metrics}

Clinical efficacy (CE) metrics capture how semantically coherent the generated and corresponding reference reports are w.r.t. an array of prominent clinical observations. To ensure comparability of results, we specifically follow \cite{miura2021improving} in calculating the CE scores micro averaged over five observations, and \cite{nicolson2022improving} in calculating the CE scores example-based averaged over all 14 observations as follows: CheXbert \cite{smit2020combining} - a BERT\cite{devlin2019bert}-based information extraction system - is first used to classify the status of 14 observations as either \emph{positive}, \emph{negative}, \emph{uncertain}, or \emph{no mention} for each generated report and corresponding reference report. The observations consist of 12 types of diseases as well as \emph{"Support Devices"} and \emph{"No Finding"}. Next, these multi-class classifications are converted to binary-class. \cite{nicolson2022improving} performs this conversion by considering \emph{positive} as the positive class, and \emph{negative}, \emph{uncertain} and \emph{no mention} as the negative class. In contrast, \cite{miura2021improving} considers \emph{positive} and \emph{uncertain} the positive class, and \emph{negative} and \emph{no mention} the negative class. Finally, \cite{nicolson2022improving} calculates the example-based precision, recall, and F1 scores over all 14 observations by comparing the classifications for each generated report and corresponding reference report. In contrast, \cite{miura2021improving} calculates the micro average precision, recall, and F1 scores over a subset of 5 observations: \emph{atelectasis}, \emph{cardiomegaly}, \emph{consolidation}, \emph{edema}, and \emph{pleural effusion}. We follow each approach respectively when comparing our results with the two works.

\subsection{Variation sampling for evaluation of selection-based sentence generation}\label{appendix:variation_sampling}
The variation sampling experiments for the evaluation of selection-based sentence generation, as showcased in Fig. 4 and with results shown in Fig. 5 from the main paper, were conducted as follows.
First, we select the first 1000 samples from the test set to reduce the required computational resources. We then use our (trained) RGRG model for selection-based sentence generation inference (see the third paragraph of Sec. 3.4 in the main paper) on this subset. Instead of letting radiologists manually draw bounding boxes, we randomly modify the ground-truth bounding boxes from those samples and use them during inference (\ie, pass them through RoI pooling). We investigate three types of variations independently: position, aspect ratio, and scale of the bounding boxes. For each of these cases, we run several experiments with different degrees of random variations.
For a specific type of variation (\ie, position, aspect ratio, or scale) and a degree of variation as defined by the 1-$\sigma$ interval (\ie, one standard deviation, as used in the $x$-axis of Fig. 5 in the main paper), an experiment corresponds to a single inference pass through all of the 1000 samples.
 We compute the micro-averaged per-anatomy METEOR score for each experiment and compare it to the default case without any variations, \ie inference on the 1000 samples using the ground-truth bounding boxes.

In a single experiment, we sample the variation for each anatomical region in each sample independently.
Assume the ground-truth box for an anatomical region is defined by its upper left $(x_1, y_1)$ and lower right $(x_2, y_2)$ corners and has width $w = x_2 - x_1$ and height $h = y_2 - y_1$. 
We sample the (additive) \textbf{position variations} $\Delta x \in (-\infty, +\infty)$ and $\Delta y \in (-\infty, +\infty)$ for the given standard deviation $\sigma$ from a zero-mean normal distribution $\mathcal{N}$ as
\begin{align}
    &\Delta x \sim \mathcal{N}(0, \sigma^2) \,, &\Delta y \sim \mathcal{N}(0, \sigma^2) \,,
\end{align}
and then compute the modified box $(\hat{x_1}, \hat{y_1}, \hat{x_2}, \hat{y_2})$ by varying the original box additively in relation to its size as
\begin{equation}
\begin{aligned}
    &\hat{x_1} = x_1 + \Delta x \cdot w\,, &\hat{x_2} = x_2 + \Delta x \cdot w \,,\\
    &\hat{y_1} = y_1 + \Delta y \cdot h\,, &\hat{y_2} = y_2 + \Delta y \cdot h \,.
\end{aligned}
\end{equation}

Aspect ratio and scale are varied multiplicatively and we, therefore, sample from the normal distribution in log-space (of aspect ratio or scale variations). In other words, the variations are Lognormal distributed. The (multiplicative) \textbf{aspect ratio} variation $\Delta a \in (0, +\infty)$ is sampled as
\begin{align}
    \Delta a \sim \text{Lognormal}(0, \sigma^2)\,,
\end{align}
i.e.\
\begin{align}
    \ln(\Delta a) \sim \mathcal{N}(0, \sigma^2)\,,
\end{align}
and similarly, the (multiplicative) \textbf{scale variation} $\Delta s \in (0, +\infty)$ is sampled as
\begin{align}
    \Delta s \sim \text{Lognormal}(0, \sigma^2)\,.
\end{align}
The 1-$\sigma$ interval for both cases is therefore defined as $[e^{-\sigma}, e^\sigma]$.
Given a sampled aspect ratio variation $\Delta a$ and a ground-truth box $(x_1, y_1, x_2, y_2)$ with aspect ratio $a = \frac{w}{h}$ and area $A = w \cdot h$, we first compute the modified aspect ratio $\hat{a}$ as
\begin{align}
    \hat{a} = \Delta a \cdot a \,,
\end{align}
then compute the modified width $\hat{w}$ and height $\hat{h}$ using the unmodified area $A$ as
\begin{equation}
\begin{aligned}
    &\hat{w} = \sqrt{A \cdot \hat{a}}\,, 
    &\hat{h} = \sqrt{\frac{A}{\hat{a}}} \,,
\end{aligned}
\end{equation}
to finally compute the modified box as
\begin{equation}
\begin{aligned}
\label{eq:coords_from_sizes}
    &\hat{x_1} = x_1 + \frac{w - \hat{w}}{2}
    &\hat{x_2} = x_2 - \frac{w - \hat{w}}{2}\,,\\
    &\hat{y_1} = y_1 + \frac{h - \hat{h}}{2}
    &\hat{y_2} = y_2 - \frac{h - \hat{h}}{2} \,.
\end{aligned}
\end{equation}
Similarly, given a sampled scale variation $\Delta s$, we first compute the updated width and height  as
\begin{align}
    &\hat{w} = \Delta s \cdot w\,, 
    &\hat{h} = \Delta s \cdot h\,,
\end{align}
and then again use \eqref{eq:coords_from_sizes} to compute the modified box.

\end{document}